\documentclass[10pt,journal,compsoc]{IEEEtran}

\ifCLASSOPTIONcompsoc
  \usepackage[nocompress]{cite}
\else
  \usepackage{cite}
\fi

\ifCLASSINFOpdf
\else
\fi

\usepackage{multirow}
\usepackage{xcolor}
\usepackage{graphicx}
\usepackage{amsmath}
\usepackage{amssymb}
\usepackage{wrapfig}
\usepackage{booktabs}
\usepackage{color}
\usepackage{verbatim}
\usepackage{caption}
\usepackage{ulem}
\usepackage[pagebackref=false,breaklinks=true,colorlinks,bookmarks=false,urlcolor=blue]{hyperref}
\usepackage{epsfig}
\usepackage{enumitem}
\usepackage{placeins}
\usepackage{ragged2e}

\usepackage{bm}
\usepackage{soul}
\usepackage{makecell}

\newcommand{\ie}{\textit{i}.\textit{e}., }

\DeclareRobustCommand{\rev}[1]{{\sethlcolor{white}\hl{#1}}}

\DeclareRobustCommand{\revminor}[1]{\sethlcolor{white}\hl{#1}}

\newcommand{\bo}[1]{\textcolor{black}{#1}}
\newcommand{\qy}[1]{\textcolor{black}{#1}}

\newcommand{\nickname}{RandLA-Net}

\newcommand{\mathcolorbox}[2]{\colorbox{#1}{$\displaystyle #2$}}

\begin{document}


\title{Learning Semantic Segmentation of Large-Scale Point Clouds with Random Sampling}

\author{Qingyong Hu, Bo Yang\textsuperscript{$\ast$},
        Linhai Xie, Stefano Rosa, Yulan Guo,
        \\ Zhihua Wang, Niki Trigoni and Andrew Markham
\IEEEcompsocitemizethanks{\IEEEcompsocthanksitem Q. Hu, L. Xie, S. Rosa, Z. Wang, N. Trigoni and A. Markham are with the Department of Computer Science, University of Oxford, United Kingdom. B. Yang is with the Department of Computing, The Hong Kong Polytechnic University, HKSAR. Y. Guo is with the School of Electronics and Communication Engineering, Sun Yat-sen University, and the College of Electronic Science and Technology, National University of Defense Technology, China. 
\IEEEcompsocthanksitem Corresponding author: Bo Yang (bo.yang@polyu.edu.hk).
}}

\markboth{Journal of \LaTeX\ Class Files}
{Guo \MakeLowercase{\textit{et al.}}: Deep Learning for 3D Point Clouds: A Survey}

\IEEEtitleabstractindextext{%
\begin{abstract}
\justifying
We study the problem of efficient semantic segmentation of large-scale 3D point clouds. By relying on expensive sampling techniques or computationally heavy pre/post-processing steps, most existing approaches are only able to be trained and operate over small-scale point clouds. In this paper, we introduce \textbf{\nickname{}}, an efficient and lightweight neural architecture to directly infer per-point semantics for large-scale point clouds. The key to our approach is to use random point sampling instead of more complex point selection approaches. Although remarkably computation and memory efficient, random sampling can discard key features by chance. To overcome this, we introduce a novel local feature aggregation module to progressively increase the receptive field for each 3D point, thereby effectively preserving geometric details. Comparative experiments show that our \nickname{} can process 1 million points in a single pass up to 200$\times$ faster than existing approaches. Moreover, extensive experiments on five large-scale point cloud datasets, including Semantic3D, SemanticKITTI, Toronto3D, NPM3D  and S3DIS, demonstrate the state-of-the-art semantic segmentation performance of our \nickname{}.

\end{abstract}

}

\maketitle

\IEEEdisplaynontitleabstractindextext

%
\IEEEpeerreviewmaketitle

\IEEEraisesectionheading{\section{Introduction}\label{sec:introduction}}
\IEEEPARstart{E}{fficient} semantic segmentation of large-scale 3D point clouds is a fundamental and essential capability for real-time intelligent systems, such as autonomous driving and augmented reality. A key challenge is that the raw point clouds acquired by depth sensors are typically irregularly sampled, unstructured and unordered. Although deep convolutional networks show excellent performance in structured 2D computer vision tasks, they cannot be directly applied to this type of unstructured data.

\begin{figure}[t]
\centering
\includegraphics[width=0.5\textwidth]{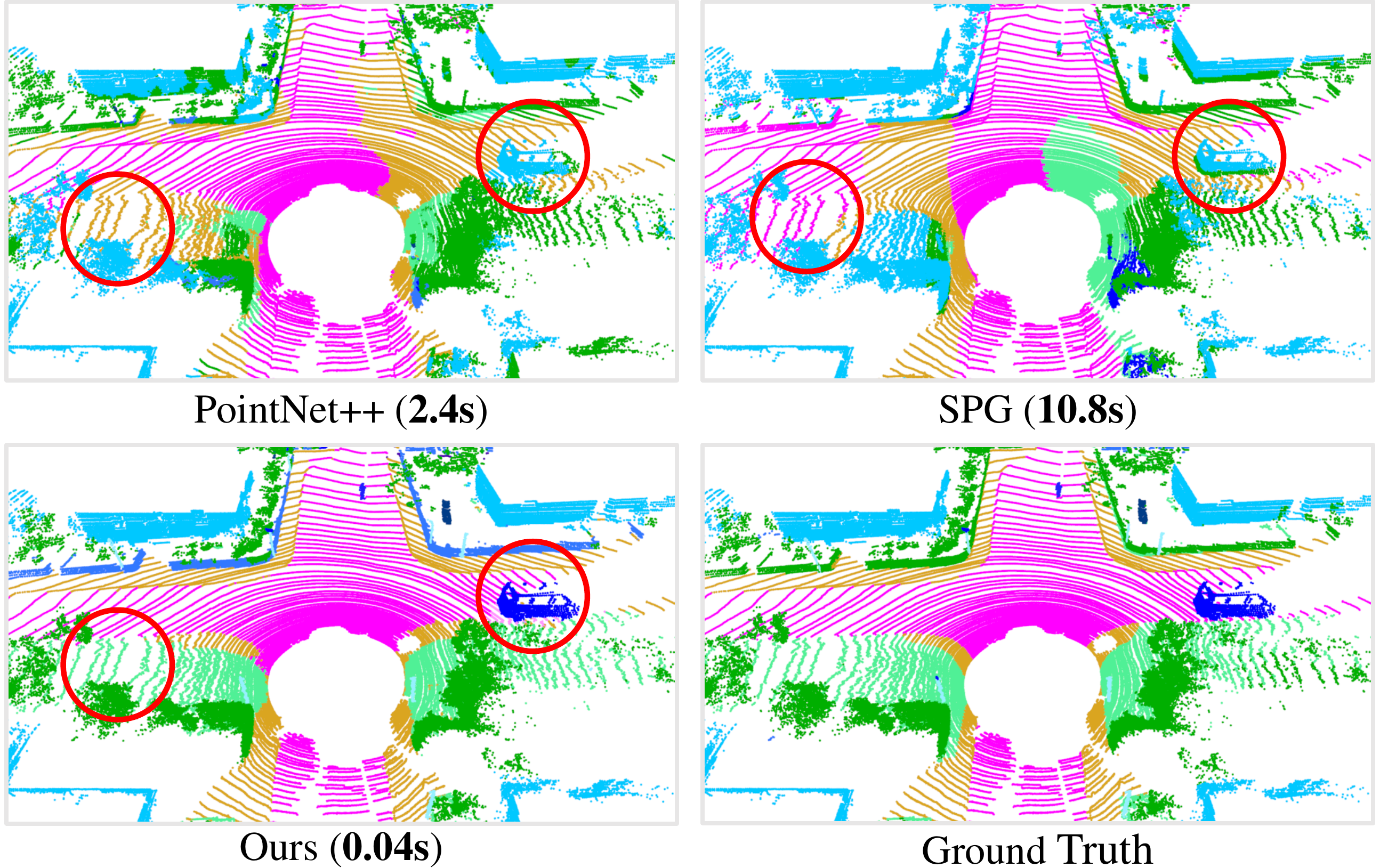}
\caption{Semantic segmentation results of PointNet++ \cite{qi2017pointnet++}, SPG \cite{landrieu2018large} and our approach on SemanticKITTI \cite{behley2019semantickitti}. Our \nickname{} takes only 0.04s to directly process a large point cloud with 81920 points over 150$\times$130$\times$10 $m^3$ in 3D space, which is up to 200$\times$ faster than SPG. Red circles highlight the superior segmentation accuracy of our approach. \label{fig:illustration}}
\end{figure}

Recently, the pioneering work PointNet \cite{qi2017pointnet} has emerged as a promising approach for directly processing 3D point clouds. It learns per-point features using shared multilayer perceptrons (MLPs). This is computationally efficient but fails to capture wider context information for each point. To learn richer local structures, many dedicated neural modules have been subsequently and rapidly introduced. These modules can be generally categorized as: 1) neighbouring feature pooling \cite{qi2017pointnet++, so-net, RSNet, pointweb, zhang2019shellnet}, 2) graph message passing \cite{dgcnn, KCNet,local_spectral,GACNet, clusternet, HPEIN, Agglomeration, simonovsky2017dynamic}, 3) kernel-based convolution \cite{su2018splatnet, hua2018pointwise, wu2018pointconv, octree_guided, ACNN, Geo-CNN, thomas2019kpconv, mao2019interpolated, wang2018deep, ummenhofer2019lagrangian, xiong2019deformable}, and 4) attention-based aggregation \cite{xie2018attentional, PCAN, Yang2019ModelingPC, AttentionalPointNet}. Although these approaches achieve impressive results for object recognition and semantic segmentation, most of them are limited to extremely small 3D point clouds (e.g., 4k points or  1$\times$1 meter blocks) and cannot be directly extended to larger point clouds (e.g., millions of points and up to 200$\times$200 meters) without preprocessing steps such as block partition. The reasons for this limitation are three-fold. 1) The commonly used point-sampling methods of these networks are either computationally expensive or memory inefficient. For example, the widely employed farthest-point sampling \cite{qi2017pointnet++} takes over 200 seconds to sample 10\% of 1 million points. 2) Most existing local feature learners usually rely on computationally expensive kernelisation or graph construction, thereby being unable to process massive number of points. 3) For a large-scale point cloud, which usually consists of hundreds of objects, the existing local feature learners are either incapable of capturing complex structures, or do so inefficiently, due to their limited size of receptive fields.

A handful of recent works have started to tackle the task of directly processing large-scale point clouds. SPG \cite{landrieu2018large} preprocesses the large point clouds as super graphs before applying neural networks to learn per super-point semantics. However, the preprocessing steps are too computationally heavy to be deployed in real-time applications. Both FCPN \cite{rethage2018fully} and PCT \cite{PCT} combine voxelization and point-level networks to process massive point clouds. However, they still partition the point clouds into small blocks for learning, resulting in the overall performance being sub-optimal.

In this paper, we aim to design a memory and computationally efficient neural architecture, which is able to directly process large-scale 3D point clouds in a single pass, without requiring pre/post-processing steps such as voxelization, block partitioning or graph construction. However, this task is extremely challenging as it requires: 1) a memory and computationally efficient sampling approach to progressively downsample large-scale point clouds to fit in the limits of current GPUs, and 2) an effective local feature learner to progressively increase the receptive field size to preserve complex geometric structures. To this end, we first systematically demonstrate that \textbf{random sampling} is a key enabler for deep neural networks to efficiently process large-scale point clouds. However, random sampling can discard key information, especially for objects with sparse points.
To counter the potentially detrimental impact of random sampling, we propose a new and efficient \textbf{local feature aggregation module} to capture complex local structures over progressively smaller point-sets.

Amongst existing sampling methods, farthest point sampling and inverse density sampling are the most frequently used for small-scale point clouds \cite{qi2017pointnet++, wu2018pointconv, li2018pointcnn, pointweb, Groh2018flexconv}. As point sampling is such a fundamental step within these networks, we investigate the relative merits of different approaches in Section \ref{Sub-sampling}, 
where we see that the commonly used sampling methods limit scaling towards large point clouds, and act as a significant bottleneck to real-time processing. However, we identify random sampling as a more suitable strategy for large-scale point cloud processing as it is fast and scales efficiently.  Random sampling is not without cost, because prominent point features may be dropped by chance and it cannot be used directly in existing networks without incurring a performance penalty.  To overcome this issue, we design a new local feature aggregation module in Section \ref{Attentive_aggregation}, which is capable of effectively learning complex local structures by progressively increasing the receptive field size in each neural layer. In particular, for each 3D point, we firstly introduce a local spatial encoding (LocSE) unit to explicitly preserve local geometric structures. Secondly, we leverage attentive pooling to automatically keep the useful local features. Thirdly, we stack multiple LocSE units and attentive poolings as a dilated residual block, greatly increasing the effective receptive field for each point. Note that all these neural components are implemented as shared MLPs, and are therefore remarkably memory and computational efficient.

Overall, being built on the principles of simple \textbf{rand}om sampling and an effective \textbf{l}ocal feature \textbf{a}ggregator, our efficient neural architecture, named \textbf{RandLA-Net}, not only is up to 200$\times$ faster than existing approaches on large-scale point clouds, but also surpasses the state-of-the-art semantic segmentation methods on Semantic3D \cite{Semantic3D}, SemanticKITTI \cite{behley2019semantickitti} and Toronto-3D \cite{Toronto3D} benchmarks. Figure \ref{fig:illustration} shows qualitative results of our approach. Our key contributions are: 
\begin{itemize}[leftmargin=*]
    \item We analyse and compare existing sampling approaches, identifying random sampling as a suitable component for efficient learning on large-scale point clouds.
    \item We propose an effective local feature aggregation module to preserve complex local structures by progressively increasing the receptive field for each point.
    \item We demonstrate significant memory and computational gains over baselines, and surpass the state-of-the-art semantic segmentation methods on multiple large-scale benchmarks.
\end{itemize}

\bo{A preliminary version of this work has been published in 
\cite{hu2019randla} and our code is available at \textit{https://github.com/QingyongHu/RandLA-Net}}.

\section{Related Work} \label{sec:related_works}

\begin{table*}[thb]
\centering
\caption{Comparison of representative sampling methods for processing point clouds. FPS: Farthest Point Sampling; IDIS: Inverse Density Importance Sampling; PDS: Poisson Disk Sampling; RS: Random Sampling; GS: Generator-based Sampling; CRS: Continuous Relaxation based Sampling; PGS: Policy Gradient-based Sampling. Note that, \textsuperscript{1}the complexity denotes the actual computational complexity when sampling $M$ points from a large-scale point cloud $\boldsymbol{P}$ with $N$ points, $K$ denotes the number of nearest neighbours. \textsuperscript{2}The time consumption is computed for sampling 10\% points from 1 million points. We use the same hardware as in Section 3.5, unless specified otherwise.}
\label{tab:sampling}
\resizebox{\textwidth}{!}{%
\begin{tabular}{ccccl}
\hline
Type & Method & Complexity\textsuperscript{1} & Time (seconds)\textsuperscript{2} & \multicolumn{1}{c}{Description} \\ \hline
\multirow{10}{*}{\begin{tabular}[c]{@{}c@{}}Heuristic \\ Sampling\end{tabular}} & FPS \cite{qi2017pointnet++} & $\mathcal{O}(M^2N)$  & 200  & \begin{tabular}[c]{@{}l@{}} Farthest Point Sampling iteratively returns a reordering of the metric space \\ $\{p_1 \cdots p_m \cdots p_M\}$, such that each $p_m$ is the farthest point from the first $m-1$ \\points. It has good coverage of points but high computational complexity.\end{tabular} \\ \cline{2-5} 
& IDIS \cite{Groh2018flexconv} & $\mathcal{O}((K+N)logN)$ & 10 & \begin{tabular}[c]{@{}l@{}}Inverse Density Importance Sampling reorders all $N$ points according to the \\ density of each point, after which the top $M$ points are selected. The density\\ is approximated by calculating the summation of the distances between the \\ point and its nearest $K$ points.  It can control density, but sensitive to outliers \\ and noise.\end{tabular} \\ \cline{2-5} 
 & PDS \cite{bridson2007fast} & $\mathcal{O}(MN)$ & 8 & \begin{tabular}[c]{@{}l@{}}Poisson Disk Sampling samples points from a set of $N$ points with blue noise \\ characteristics  i.e., points are sampled from a Poisson disk distribution, where \\ all samples are at least a certain distance $r$ apart.\end{tabular} \\ \cline{2-5} 
 & RS & $\mathcal{O}(M)$ & 0.004 & \begin{tabular}[c]{@{}l@{}}Random Sampling uniformly selects $M$ points from the original $N$ points, \\ each point has the same probability to be selected. It is agnostic to the total \\ number of input points, i.e., it is constant-time and hence is inherently scalable.\end{tabular} \\ \hline
\multirow{9}{*}{\begin{tabular}[c]{@{}c@{}}Learning-based \\ Sampling\end{tabular}} & GS \cite{learning2sample} & - & 1200 & \begin{tabular}[c]{@{}l@{}}Generator based Sampling learns to generate a small set of points to \\ approximately represent  the original point set. FPS matching is required \\ during inference.\end{tabular} \\ \cline{2-5} 
 & CRS \cite{concrete} & \multicolumn{2}{c}{\begin{tabular}[c]{@{}c@{}} Estimated 300 GB GPU\\ memory footprint\end{tabular}} & \begin{tabular}[c]{@{}l@{}}Continuous Relaxation based Sampling uses the reparameterization trick \\ to relax the sampling operation to a continuous domain for end-to-end \\ training. A large weight matrix is required when sampling all the new \\ points simultaneously with one pass matrix multiplication, leading to an \\ unaffordable memory cost.\end{tabular} \\ \cline{2-5} 
 & PGS \cite{show_attend} & \multicolumn{2}{c}{\begin{tabular}[c]{@{}c@{}}$\mathrm{C}_{10^{6}}^{10^{5}}$\\ exploration space\end{tabular}} & \begin{tabular}[c]{@{}l@{}}Policy Gradient-based Sampling formulates the sampling operation as a \\ Markov decision process. It sequentially learns a probability distribution\\ to sample the points. However, the learned probability has high variance \\ due to the extremely large exploration space when the point cloud is large.\end{tabular} \\ \hline
\end{tabular}%
}
\end{table*}

To extract features from 3D point clouds, traditional approaches usually rely on hand-crafted features \cite{point_signatures, fast_hist, landrieu2017structured, hackel2016fast}. Recent learning based approaches \cite{guo2019deep, qi2017pointnet, Point_voxel_cnn} mainly include projection-based, voxel-based and point-based schemes which are outlined here.

\textbf{(1) Projection and Voxel-Based Networks.}
To leverage the success of 2D CNNs, many works \cite{li2016vehicle_rss, chen2017multi, PIXOR, pointpillars} project/flatten 3D point clouds onto 2D images to address the task of object detection. 
However, geometric details may be lost during  projection. For example, the commonly-used birds-eye-view projection can drop some points due to  occlusion. Alternatively, several other approaches \cite{rangenet++, salsanext, wu2018squeezeseg, wu2019squeezesegv2, xu2020squeezesegv3} use spherical projection to convert the LiDAR point clouds back to raw range images without dropping points. However, these methods usually suffer from blurry CNN outputs and quantization errors in practice \cite{rangenet++}.
Point clouds can also be voxelized into dense 3D grids and then processed with powerful 3D CNNs \cite{ pointgrid, vvnet, Fast_point_rcnn}. However, their memory consumption and computational time are significant if a high voxel grid resolution is required. This issue can be alleviated using Octree \cite{riegler2017octnet} or sparse-tensor \cite{sparse, 4dMinkpwski}, but these advanced data structures and sophisticated operations are not always easily supported by existing tools, especially on GPUs.

\textbf{(2) Point Based Networks.}
Inspired by PointNet/PointNet++ \cite{qi2017pointnet, qi2017pointnet++}, many recent works introduced sophisticated neural modules to learn per-point local features. These modules can be generally classified as 1) neighbouring feature pooling \cite{so-net, RSNet, pointweb, zhang2019shellnet}, 2) graph message passing \cite{dgcnn, KCNet,local_spectral,GACNet, clusternet, HPEIN, Agglomeration, Li_2019_ICCV}, 3) kernel-based convolution \cite{su2018splatnet, hua2018pointwise, wu2018pointconv, octree_guided, ACNN, Geo-CNN, thomas2019kpconv, mao2019interpolated, rosu2019latticenet}, and 4) attention-based aggregation \cite{xie2018attentional, PCAN, Yang2019ModelingPC, AttentionalPointNet}. Although these networks have shown promising results on small point clouds, most of them cannot directly scale up to large scenarios due to their high computational and memory costs. Compared with them, our proposed \nickname{} is distinguished in three ways: 1) it only relies on random sampling within the network, thereby requiring much less memory and computation; 2) the proposed local feature aggregator can obtain successively larger receptive fields by explicitly considering the local spatial relationship and point features, thus being more effective and robust for learning complex local patterns; 3) the entire network only consists of shared MLPs coupled with the simple random sampling, therefore being efficient for large-scale point clouds. 

\textbf{(3) Learning for Large-scale Point Clouds}.
SPG \cite{landrieu2018large} preprocesses the large point clouds as superpoint graphs to learn per superpoint semantics. However, both the geometric partition and superpoint graph construction are computationally expensive.
The recent FCPN \cite{rethage2018fully} and PCT \cite{PCT} apply both voxel-based and point-based networks to process the massive point clouds. Based on the assumption that points are sampled from locally Euclidean surfaces, \bo{TangentConv \cite{tangentconv} firstly} projects the local surface on the tangent plane and then operates on the projected geometry. \bo{Despite being able to process} 
large-scale point clouds, it requires a relatively heavy  preprocessing step to calculate the normal. In contrast, our \nickname{} is end-to-end trainable without requiring additional expensive operations.

\textbf{(4) Sampling Methods for 3D Point Clouds}. Existing point sampling approaches \cite{qi2017pointnet++, li2018pointcnn, Groh2018flexconv, learning2sample, concrete, wu2018pointconv} can be roughly classified into heuristic and learning-based methods. Farthest point sampling \cite{qi2017pointnet++, li2018pointcnn} is the most commonly used heuristic sampling strategy in recent works. It iteratively samples the  points most distant to the remaining subset from the entire point set. Groh et al. \cite{Groh2018flexconv} use inverse density importance sub-sampling (IDIS) to preserve points with lower density. Hermosilla et al. \cite{pedro} utilize Poisson disk sampling (PDS) to achieve a uniform distribution for the sampled points. For learning-based approaches, Yang et al. \cite{Yang2019ModelingPC} introduce Gumbel subset sampling, an end-to-end learnable and task-agnostic sampling method, to obtain better performance for downstream tasks. Dovrat et al \cite{learning2sample} propose a generater network to directly generate a point set to approximate the original point set. However, all these methods are either ineffective or computationally heavy as evaluated in Section \ref{Sub-sampling} and Table \ref{tab:sampling}.

\section{Proposed Methods}

\subsection{Overview}

\begin{figure}[htb]
\centering
\includegraphics[width=0.5\textwidth]{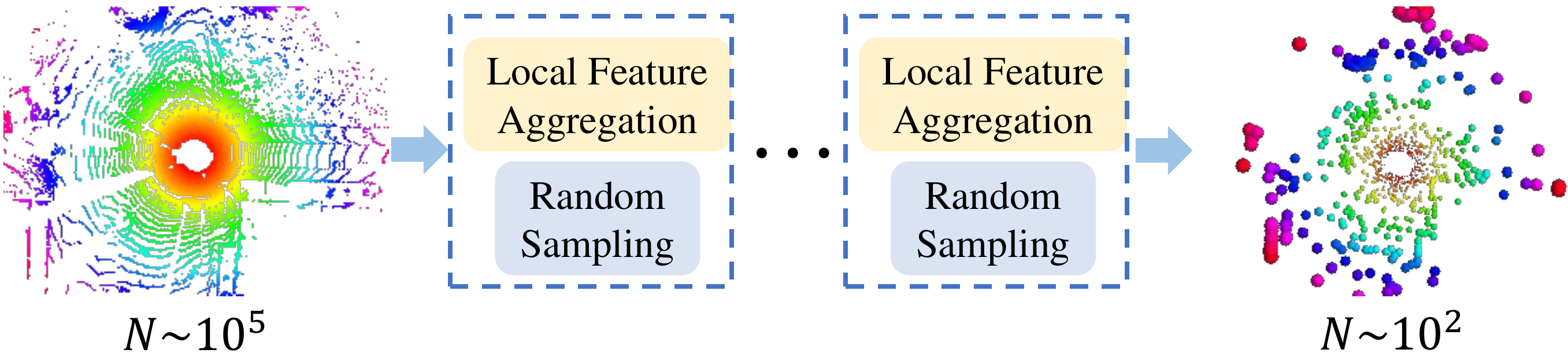}
\caption{In each layer of \nickname{}, the large-scale point cloud is significantly downsampled, yet is capable of retaining features necessary for accurate segmentation.}
\label{fig:sampling}
\end{figure}

As illustrated in Figure \ref{fig:sampling}, \rev{given a large-scale point cloud with millions of points which could span hundreds of meters, to process it with a deep neural network inevitably requires those points to be progressively and efficiently downsampled in each neural layer, without losing the useful point features.} In our \nickname{}, we propose to use the simple and fast approach of random sampling to greatly decrease point density, whilst applying a carefully designed local feature aggregator to retain prominent features. This allows the entire network to achieve an excellent trade-off between efficiency and effectiveness.

\begin{figure*}[thb]
\centering
\includegraphics[width=\textwidth]{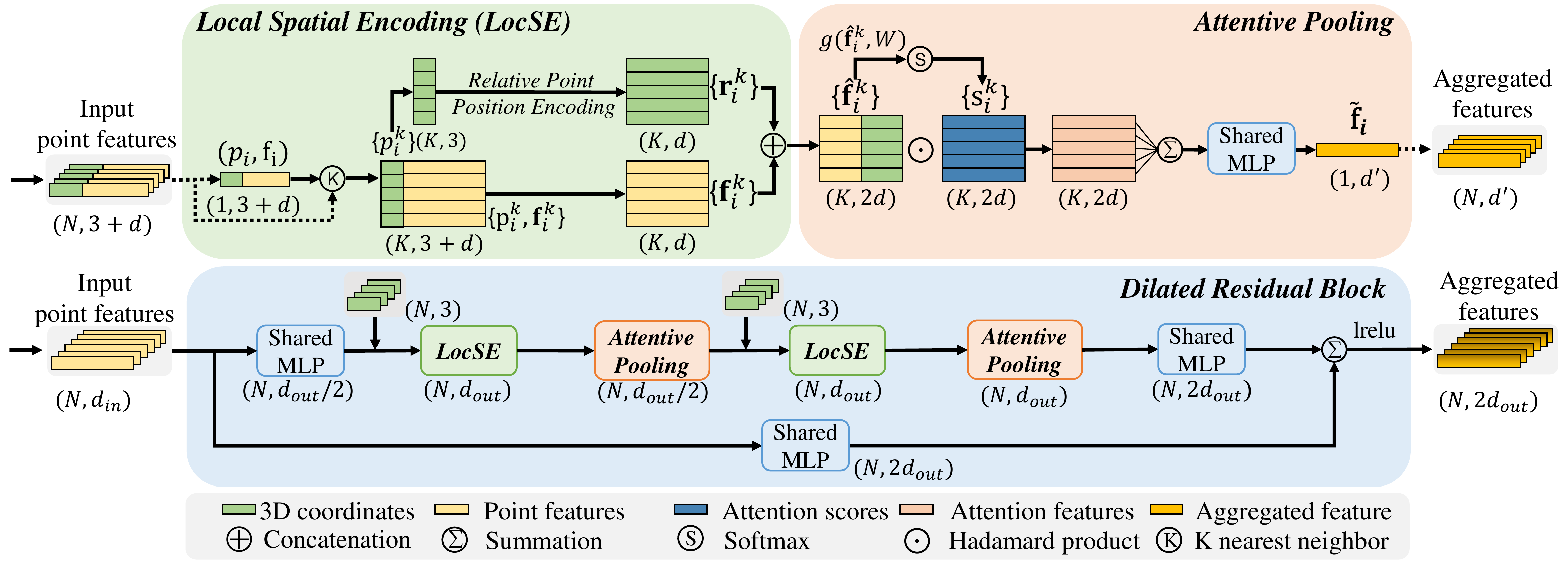}
\caption{The proposed local feature aggregation module. The top panel shows the local spatial encoding block that extracts features, and the attentive pooling mechanism that weights the important neighbouring features, based on the local context and geometry. The bottom panel shows how two of these components are chained together, to increase the receptive field size, within a residual block.}
\label{fig:network}
\end{figure*}

\subsection{The quest for efficient sampling}
\label{Sub-sampling}

To determine the most suitable method for processing large-scale point clouds, we provide an empirical analysis and comparison of different sampling approaches. In particular, the relative merits and complexity of the commonly used sampling strategies are shown in Table \ref{tab:sampling}.

Overall, FPS, IDIS and GS are too computationally expensive to be applied for large-scale point clouds. CRS approaches have an excessive memory footprint and PGS is hard to learn. PDS is relatively faster, but has worse performance, as shown in the ablation study. By contrast, random sampling has the following two advantages: 1) it is remarkably computational efficient as it is agnostic to the total number of input points, 2) it does not require extra memory for computation. Therefore, we believe that random sampling is more suitable than other approaches to efficiently process large-scale point clouds. However, random sampling may result in many useful point features being dropped. To overcome it, we propose a powerful local feature aggregation module as presented in the next section.

\subsection{Local Feature Aggregation}
\label{Attentive_aggregation}
As shown in Figure \ref{fig:network}, our local feature aggregation module is applied to each 3D point in parallel and it consists of three neural units: 1) local spatial encoding (LocSE), 2) attentive pooling, and 3) dilated residual block.

\textbf{(1) Local Spatial Encoding}\\
Given a point cloud $\boldsymbol{P}$ together with per-point features (e.g., raw RGB, or intermediate learned features), this local spatial encoding unit explicitly embeds the x-y-z coordinates of all neighbouring points, such that the corresponding point features are always aware of their relative spatial locations. This allows the LocSE unit to explicitly observe the local geometric patterns, thus eventually benefiting the entire network to effectively learn complex local structures. In particular, this unit includes the following steps:

\textit{Finding Neighbouring Points.} For the $i^{th}$ point, its neighbouring points are firstly gathered by the simple $K$-nearest neighbours (KNN) algorithm for efficiency. The KNN is based on point-wise Euclidean distances.  

\textit{Relative Point Position Encoding.} For each of the nearest $K$ points $\{p_i^1 \cdots p_i^k \cdots p_i^K\}$ of the center point $p_i$, we explicitly encode the relative point position as follows:
\begin{equation}
  \mathbf{r}_{i}^{k} = MLP\Big(p_i \oplus p_i^k \oplus (p_i-p_i^k) \oplus ||p_i-p_i^k||\Big)
\label{Eq1}
\end{equation}
where $p_i$ and $p_i^k$ are the absolute x-y-z positions of points, $\oplus$ is the concatenation operation, and $||\cdot||$ calculates the Euclidean distance between the neighbouring and center points. It seems that $\mathbf{r}_{i}^{k}$ is encoded from redundant point positions. Interestingly, this tends to aid the network to learn local features and obtains good performance in practice.

\textit{Point Feature Augmentation.} For each neighbouring point $p_i^k$, the encoded relative point positions $\mathbf{r}_{i}^{k}$ are concatenated with its corresponding point features $\mathbf{f}_i^k$, obtaining an augmented feature vector $\mathbf{\hat{f}}_i^k$. For simplicity, we keep the feature vectors $\mathbf{r}_{i}^{k}$ and $\mathbf{f}_i^k$ with the same dimension in the implementation, but they are flexible and can have different dimensions.

Eventually, the output of the LocSE unit is a new set of neighbouring features $\mathbf{\hat{F}}_i = \{\mathbf{\hat{f}}_i^1 \cdots \mathbf{\hat{f}}_i^k \cdots \mathbf{\hat{f}}_i^K \}$, which explicitly encodes the local geometric structures for the center point $p_i$. We notice that the recent work \cite{liu2019relation} also uses point positions to improve semantic segmentation. However, the positions are used to learn point scores in \cite{liu2019relation}, while our LocSE explicitly encodes the relative positions to augment the neighbouring point features.

\textbf{(2) Attentive Pooling}\\
This neural unit is used to aggregate the set of neighbouring point features $\mathbf{\hat{F}}_i$. Existing works \cite{qi2017pointnet++, li2018pointcnn} typically use max/mean pooling to hard integrate the neighbouring features, resulting in the majority of the information being lost. By contrast, we turn to the powerful attention mechanism to automatically learn important local features. In particular, inspired by \cite{Yang_ijcv2019}, our attentive pooling unit consists of the following steps.

\begin{figure*}[t]
\centering
\includegraphics[width=1\textwidth]{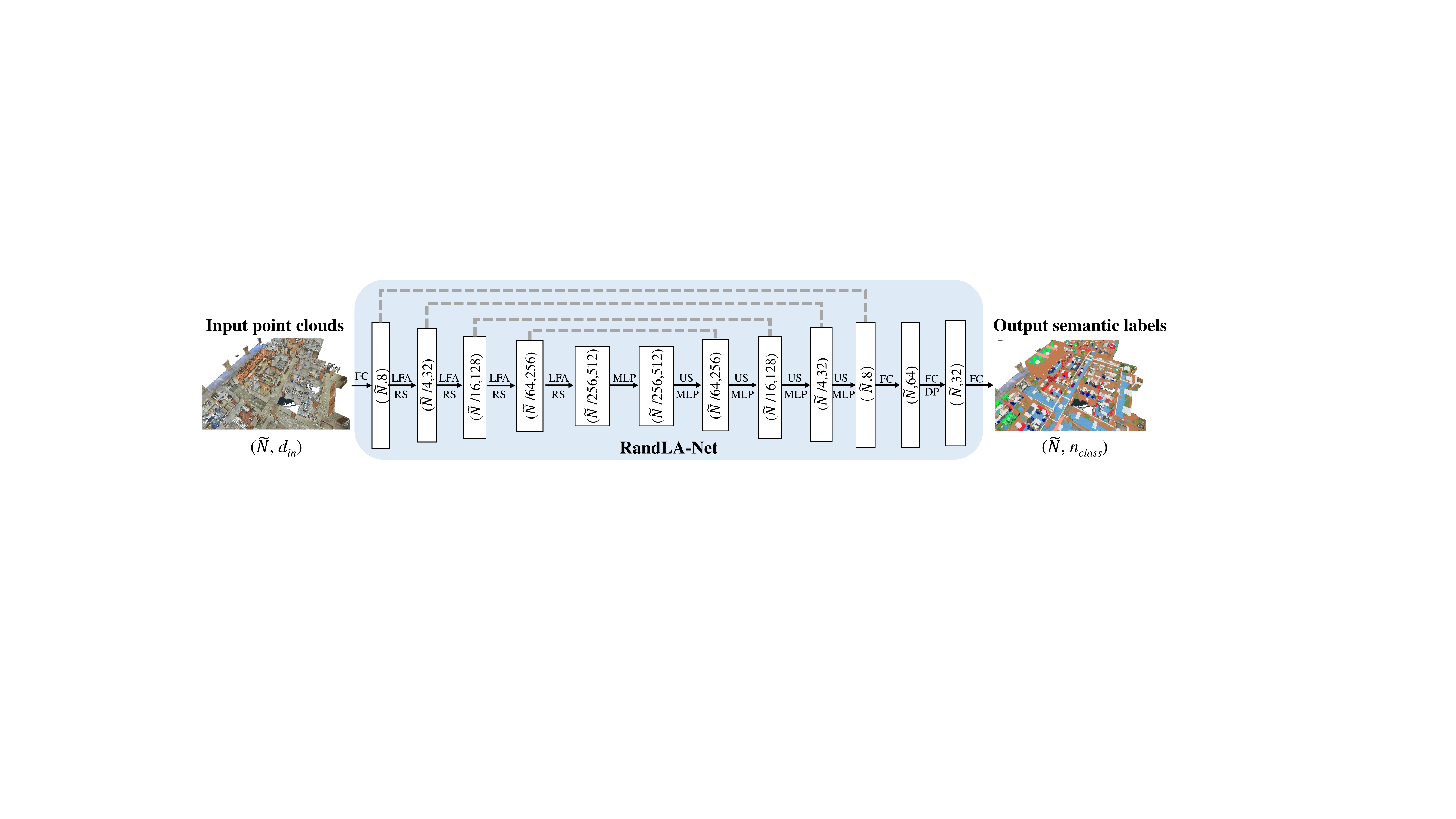}
\caption{The detailed architecture of our \nickname{}. $(\widetilde{N},D)$ represents the number of points and feature dimension respectively. FC: Fully Connected layer, LFA: Local Feature Aggregation, RS: Random Sampling, MLP: shared Multi-Layer Perceptron, US: Up-sampling, DP: Dropout.}
\label{fig:network-detailed}
\end{figure*}

\textit{Computing Attention Scores.} Given the set of local features $\mathbf{\hat{F}}_i = \{\mathbf{\hat{f}}_i^1 \cdots \mathbf{\hat{f}}_i^k \cdots \mathbf{\hat{f}}_i^K \}$, we design a shared function $g(\cdot)$ to learn a unique attention score for each feature. Essentially, the function $g(\cdot)$ consists of a shared MLP followed by $softmax$. It is formally defined as follows:
\begin{equation}
  \mathbf{s}_{i}^{k} = g(\mathbf{\hat{f}}_i^k, \boldsymbol{W})
  \label{Eq2}
\end{equation}
where $\boldsymbol{W}$ is the learnable weights of a shared MLP.

\textit{Weighted Summation.} The learned attention scores can be regarded as a soft mask which automatically selects the important features. Formally, these features are weighted summed as follows:
\begin{equation}
\mathcolorbox{white}{
  \mathbf{\Tilde{f}}_{i} = \sum_{k=1}^{K}(\mathbf{\hat{f}}_i^k \odot   \mathbf{s}_{i}^{k})}
\end{equation}
where $\odot$ is the element-wise product. To summarize, given the input point cloud $\boldsymbol{P}$, for the $i^{th}$ point $p_i$, our LocSE and Attentive Pooling units learn to aggregate the geometric patterns and features of its $K$ nearest points, and finally generate an informative feature vector $\mathbf{\Tilde{f}}_{i}$.

\textbf{(3) Dilated Residual Block}\\
Since the large point clouds are going to be substantially downsampled, it is desirable to significantly increase the receptive field for each point, such that the geometric details of input point clouds are more likely to be preserved, even if some points are dropped. As shown in Figure \ref{fig:network}, inspired by the successful ResNet \cite{he2016deep} and the effective dilated networks \cite{DPC}, we stack multiple LocSE and Attentive Pooling units to achieve the dilation of point receptive fields, and add a skip connection to achieve residual learning.

\begin{figure}[t]
\centering
\includegraphics[width=0.5\textwidth]{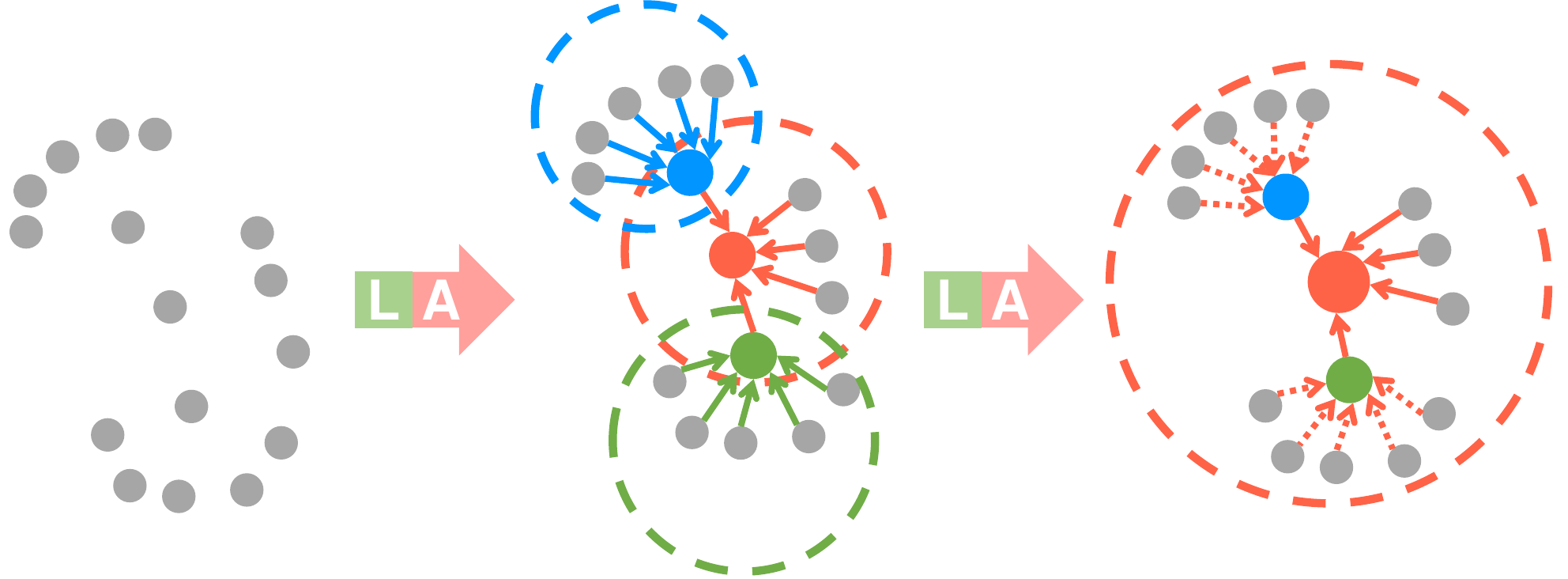}
\caption{Illustration of the dilated residual block which significantly increases the receptive field (dotted circle) of each point, colored points represent the aggregated features. L: Local spatial encoding, A: Attentive pooling.}
\label{fig:Residual}
\end{figure}

To further illustrate the capability of our dilated residual block, Figure \ref{fig:Residual} shows that the red 3D point observes $K$ neighbouring points after the first LocSE/Attentive Pooling operation, and then is able to receive information from up to $K^2$ neighbouring points i.e. its two-hop neighbourhood after the second. 
This is a cheap way of dilating the receptive field and expanding the effective neighbourhood through feature propagation. Theoretically, the more units we stack, the more powerful this block as its sphere of reach becomes greater and greater. However, more units would inevitably sacrifice the overall computation efficiency. In addition, the entire network is likely to be over-fitting. In our \nickname{}, we simply stack two sets of LocSE and Attentive Pooling as the standard residual block, achieving a satisfactory balance between efficiency and effectiveness. 

Overall, our local feature aggregation module is designed to effectively preserve complex local structures via explicitly considering neighbouring geometries and significantly increasing receptive fields. Moreover, this module only consists of feed-forward MLPs, thus being computationally efficient.   

\subsection{Network Architecture}
\label{network_structure}
Figure \ref{fig:network-detailed} shows the detailed architecture of \nickname{}, \bo{which stacks multiple local feature aggregation modules and random sampling layers.} The network follows the widely-used encoder-decoder architecture with skip connections. The input point cloud is first fed to a shared MLP layer to extract per-point features. Four encoding and decoding layers are then used to learn features for each point. At last, three fully-connected layers and a dropout layer are used to predict the semantic label of each point. Note that, our \nickname{} can be wider and deeper by altering the number of layers, feature channels, and adjusting the sampling rate. However, larger models require additional computation and may more easily lead to overfitting. The details of each component are as follows: \\

\noindent\textbf{Network Input:} The input is a large-scale point cloud with a size of $\widetilde{N}\times d_{in}$ (the batch dimension is dropped for simplicity), where $\widetilde{N}$ is the total number of input points, $d_{in}$ is the feature dimension of each input point. For both S3DIS \cite{2D-3D-S} and Semantic3D \cite{Semantic3D} datasets, each point is represented by its 3D coordinates and color information (i.e., x-y-z-R-G-B), while each point of the SemanticKITTI \cite{behley2019semantickitti} dataset is only represented by 3D coordinates.
\\

\noindent\textbf{Encoding Layers:} Four encoding layers are used in our network to progressively reduce the size of the point clouds and increase the per-point feature dimensions. Each encoding layer consists of a local feature aggregation module (Section \ref{Attentive_aggregation}) and a random sampling operation (Section \ref{Sub-sampling}). The point cloud is downsampled with a four-fold decimation ratio. In particular, only 25\% of the point features are retained after each layer, i.e.,  $(\widetilde{N}\rightarrow \frac{\widetilde{N}}{4}\rightarrow \frac{\widetilde{N}}{16}\rightarrow \frac{\widetilde{N}}{64}\rightarrow \frac{\widetilde{N}}{256})$. Meanwhile, the per-point feature dimension is gradually increased each layer to preserve more information, i.e., $(8\rightarrow32\rightarrow 128\rightarrow 256\rightarrow 512)$. \\

\noindent\textbf{Decoding Layers:} Four decoding layers are used after the encoding layers. For each layer in the decoder, we adopt the nearest-neighbor interpolation for efficiency and simplicity. In particular, the coordinates of all down-sampled points in each encoding layer are temporally stored for reference. For each query point in the decoding layers, we use the KNN algorithm to find the nearest neighboring point from the points of the previous layer. The features of the nearest point are copied to the target point.
Subsequently, the upsampled feature maps are concatenated with the intermediate feature maps produced by encoding layers through skip connections, after which a shared MLP is applied to the concatenated feature vectors.\\

\noindent\textbf{Final Semantic Prediction:} The final semantic label of each point is obtained through three shared fully-connected layers ($\widetilde{N}$, 64) $\rightarrow$ ($\widetilde{N}$, 32) $\rightarrow$ ($\widetilde{N}$, $n_{class}$) and a dropout layer. The dropout ratio is 0.5.\\

\noindent\textbf{Network Output:} The output of \nickname{} is the predicted semantics of all points, with a size of $ \widetilde{N}\times n_{class}$, where $n_{class}$ is the number of classes.

\subsection{Implementation}
\label{subsec:Implementation}

We use the same network architecture for all the five large-scale open datasets, Semantic3D \cite{Semantic3D}, SemanticKITTI \cite{behley2019semantickitti}, Toronto-3D \cite{Toronto3D}, NPM3D \cite{NPM3D}, S3DIS \cite{2D-3D-S}. The Adam optimizer \cite{kingma2014adam} with default parameters is applied. The initial learning rate is set to 0.01 and decreases by 5\% after each epoch. The network is trained for 100 epochs. We implement the KNN search in our framework based on the nanoflann\footnote{https://github.com/jlblancoc/nanoflann} package, which leverages an efficient KD-Tree data structure for fast search and query. In addition, we also use OpenMP for better parallelization, and the number of nearest points $K$ is set to 16. For several extremely large datasets such as Semantic3D \cite{Semantic3D}, which have more than $10^6$ or even $10^8$ points in a single point cloud, we crop sub-clouds to feed into our \nickname{}. During training, we sample a fixed number of points ($\sim 10^5$) from each point cloud as the input for parallelization. During testing, we iteratively infer several sub-clouds with overlaps to eventually cover all  3D points. Since many points have been inferred more than once, we follow \cite{thomas2019kpconv} to use a simple voting scheme for better performance. To alleviate the problem of class imbalance, we use weighted cross-entropy as the loss function. In particular, the weight of each class is determined by its inverse frequency in the training split. We do not use any explicit data augmentation techniques such as rotation, scaling, and translation during training. Note that, our random downsampling in each encoding layer can be regarded as implicit data augmentation. All experiments are conducted on an NVIDIA RTX2080Ti GPU.

\section{Experiments}
We first evaluate the efficiency of existing sampling approaches in Section \ref{sec:eff_sampling}, and compare the efficiency of our \nickname{} with existing networks in Section \ref{sec:eff_net}. Secondly, we conduct extensive experiments on multiple benchmarks to compare the semantic segmentation in Section \ref{subsec:sem_seg_benchmarks}. At last, our ablation study is presented in Section \ref{sec:ablation}.

\subsection{Efficiency of Random Sampling}
\label{sec:eff_sampling}

In this section, we empirically evaluate the efficiency of existing sampling approaches including FPS, IDIS, PDS, RS, GS, CRS, and PGS, which have been discussed in Section \ref{Sub-sampling}. In particular, we conduct the following 4 groups of experiments.
\begin{itemize}[leftmargin=*]
    \item Group 1. Given a small-scale point cloud ($\sim 10^3$ points), we use each sampling approach to progressively downsample it. Specifically, the point cloud is downsampled by five steps with only 25\% points being retained in each step on a single GPU i.e. a four-fold decimation ratio. This means that there are only $\sim (1/4)^5 \times 10^3$ points left in the end. This downsampling strategy emulates the procedure used in PointNet++ \cite{qi2017pointnet++}. For each sampling approach, we sum up its time and memory consumption for comparison.
    \item Group 2/3/4. The total number of points is increased towards large-scale, i.e., around $10^4, 10^5$ and $10^6$ points respectively. We use the same five sampling steps as in Group 1. 
\end{itemize}

\begin{figure}[thb]
\centering
\includegraphics[width=0.5\textwidth]{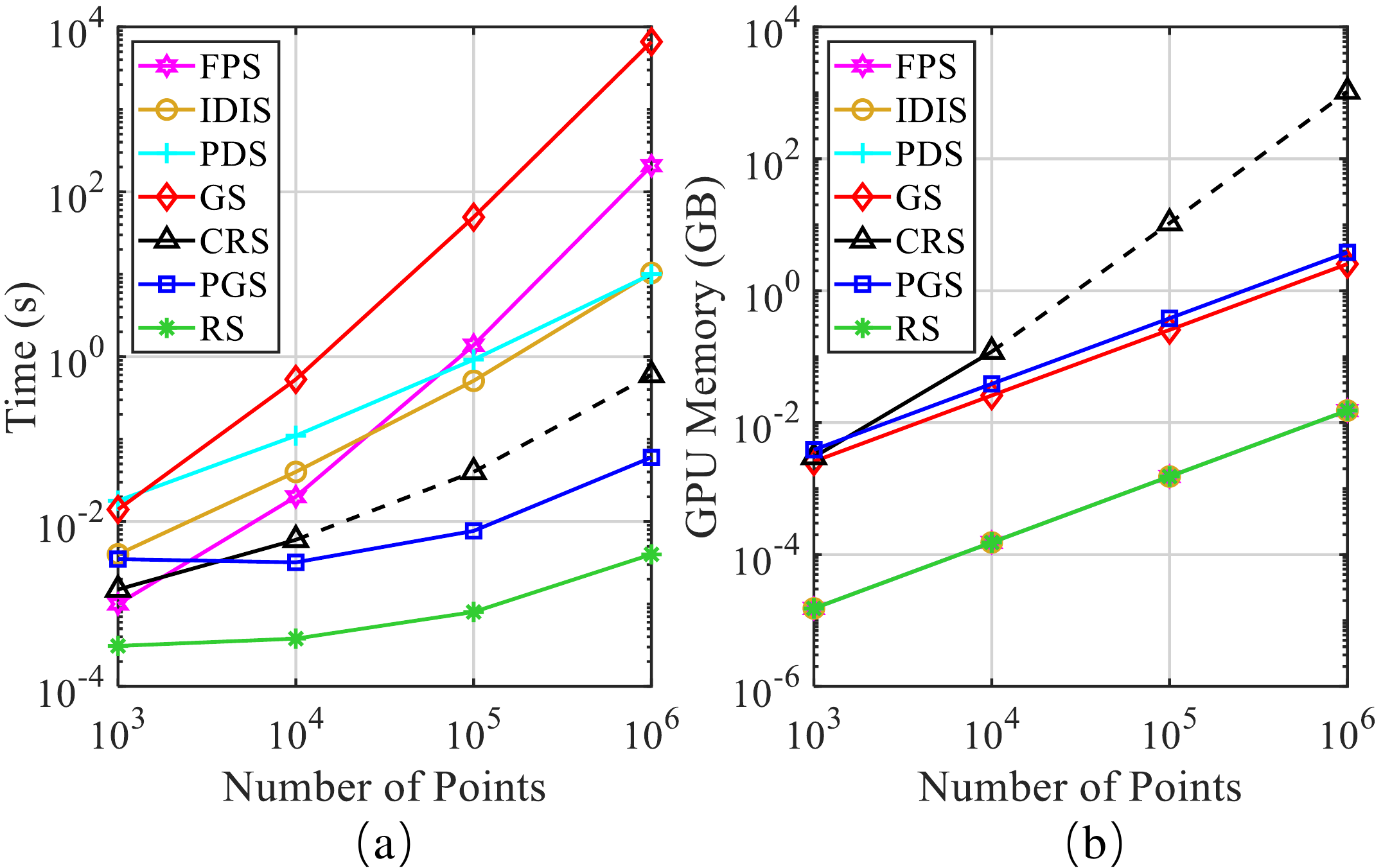}
\caption{Time and memory consumption of different sampling approaches. The dashed lines represent estimated values due to the limited GPU memory. Note that, the curves of GPU memory consumption for RPS, PDS, FPS, and IDIS overlap together, because these sampling methods mainly run on the CPU.}
\label{fig:sampling_comparison}
\end{figure}

\begin{table}[thb]
\centering
\caption{The runtime decomposition and peak memory of different approaches for semantic segmentation of input point clouds with 12$\times$81920 ($\sim 10^6$) points, which is sampled from the LiDAR point clouds on Sequence 08 (around 80K-120K points per frame) of the SemanticKITTI \cite{behley2019semantickitti} dataset. In particular, $^\dagger$this is achieved by inference half of the input point clouds twice, since PointNet is unable to process the whole inputs in a single pass. $^\ddagger$The original implementation of PointNet++ adopts an extremely inefficient KNN during trilinear interpolation. Hence, we also report the result of re-implemented PointNet++\textsuperscript{*} for a fair comparison. Note that, because SPG does not have explicit sampling and neighboring search modules, its runtime decomposition is not reported. We measure the peak memory consumption based on the TensorFlow profiler (https://www.tensorflow.org/guide/profiler). Note the peak memory may be different from the actual allocated GPU memory. }
\label{tab:efficiency}
\resizebox{0.48\textwidth}{!}{%
\begin{tabular}{rccccc}
\Xhline{2\arrayrulewidth}
\multicolumn{1}{c}{\multirow{2}{*}{}} & \multicolumn{4}{c}{Time (s)} & \multirow{2}{*}{\begin{tabular}[c]{@{}c@{}}Peak\\ Memory\end{tabular}} \\ \cline{2-5}
\multicolumn{1}{c}{} & Sampling & Neighbour. & Network. & Total &  \\ 
\Xhline{2\arrayrulewidth}
PointNet (Vanilla)$^\dagger$ \cite{qi2017pointnet} & 0 & 0 & 0.46$^\dagger$  & 0.46 & 9064Mb  \\
PointNet++ (SSG)$^\ddagger$ \cite{qi2017pointnet++}  & 1.32 & 0.32+26.98$^\ddagger$ & 0.20 & 28.82 & 4135Mb \\
PointNet++ (re-imp.)\textsuperscript{*} \cite{qi2017pointnet++}  & 1.32 & 0.52 & 0.20 & 2.04 & 4135Mb \\
SPG \cite{landrieu2018large} & -  & -  & - & 99.67 & 1663Mb \\
\textbf{RandLA-Net (Ours)} & 0.01 & 0.40 & 0.15 & 0.55 & 5198Mb \\
\Xhline{2\arrayrulewidth}
\end{tabular}%
}
\end{table}
\vspace{-0.2cm}

\textbf{Analysis.} Figure \ref{fig:sampling_comparison} compares the total time and memory consumption of each sampling approach to process different scales of point clouds. It can be seen that: 1) For small-scale point clouds ($\sim 10^3$), all sampling approaches tend to have similar time and memory consumption, and are unlikely to incur a heavy or limiting computation burden. 2) For large-scale point clouds ($\sim 10^6$), FPS/IDIS/PDS/GS/CRS/PGS are either extremely time-consuming or memory-costly for large-scale computation. By contrast, random sampling has superior time and memory efficiency overall. This result clearly demonstrates that most existing networks \cite{qi2017pointnet++, li2018pointcnn, wu2018pointconv, liu2019relation, pointweb, Yang2019ModelingPC} are only able to be optimized on small blocks of point clouds primarily because they rely on the expensive sampling approaches. Motivated by this, we use the efficient random sampling strategy in our \nickname{}.

\subsection{Efficiency of \nickname{}}
\label{sec:eff_net}
\begin{figure*}[thb]
\centering
\includegraphics[width=0.97\textwidth]{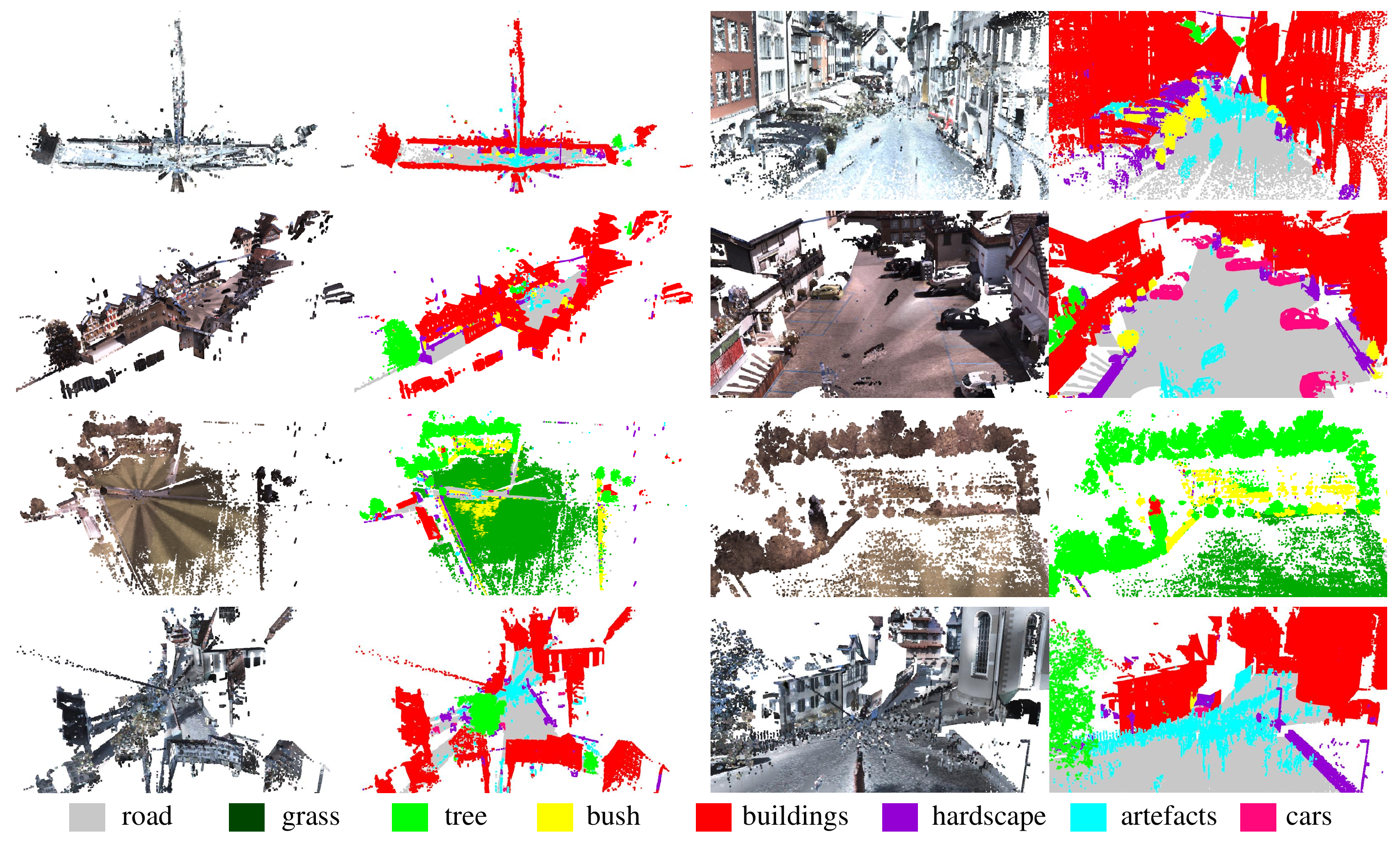}
\caption{Qualitative results of \nickname{} on the \textit{reduced-8} split of Semantic3D. From left to right: full RGB colored point clouds, predicted semantic labels of full point clouds, detailed view of colored point clouds, detailed view of predicted semantic labels. Note that the ground truth of the test set is not publicly available.}
\label{fig:reduced-8-vis}
\end{figure*}

\begin{table*}[htb]
\centering
\caption{Quantitative results of different approaches on Semantic3D (\textit{reduced-8}) \cite{Semantic3D}. This test consists of 78,699,329 points. The scores are obtained from the recent publications. Accessed on 1 May 2020.}
\label{tab:reduced-8}
\resizebox{\textwidth}{!}{%
\begin{tabular}{rcccccccccc}
\toprule[1.0pt]
Methods & mIoU(\%) & OA(\%) & man-made. & natural. & high veg. & low veg. & buildings & hard scape & scanning art. & cars \\
\toprule[1.0pt]
SnapNet\_ \cite{snapnet} & 59.1 & 88.6 & 82.0 & 77.3 & 79.7 & 22.9 & 91.1 & 18.4 & 37.3 & 64.4 \\
SEGCloud \cite{tchapmi2017segcloud} & 61.3 & 88.1 & 83.9 & 66.0 & 86.0 & 40.5 & 91.1 & 30.9 & 27.5 & 64.3 \\
RF\_MSSF \cite{RF_MSSF} & 62.7 & 90.3 & 87.6 & 80.3 & 81.8 & 36.4 & 92.2 & 24.1 & 42.6 & 56.6 \\
MSDeepVoxNet \cite{msdeepvoxnet} & 65.3 & 88.4 & 83.0 & 67.2 & 83.8 & 36.7 & 92.4 & 31.3 & 50.0 & 78.2 \\
ShellNet \cite{zhang2019shellnet} & 69.3 & 93.2 & 96.3 & 90.4 & 83.9 & 41.0 & 94.2 & 34.7 & 43.9 & 70.2 \\
GACNet \cite{GACNet} & 70.8 & 91.9 & 86.4 & 77.7 & \textbf{88.5} & \textbf{60.6} & 94.2 & 37.3 & 43.5 & 77.8 \\
SPG \cite{landrieu2018large} & 73.2 & 94.0 & 97.4 & 92.6 & 87.9 & 44.0 & 83.2 & 31.0 & 63.5 & 76.2 \\
KPConv \cite{thomas2019kpconv} & 74.6 & 92.9 & 90.9 & 82.2 & 84.2 & 47.9 & 94.9 & 40.0 & \textbf{77.3} & \textbf{79.7} \\
RGNet \cite{RGNet} & 74.7 & 94.5	& \textbf{97.5}	& \textbf{93.0} & 88.1 & 48.1 & 94.6 & 36.2 & 72.0 & 68.0  \\
\textbf{\nickname{} (Ours)} & \textbf{77.4} & \textbf{94.8} & 95.6 & 91.4 & 86.6 & 51.5 & \textbf{95.7} & \textbf{51.5} & 69.8 & 76.8
\\
\bottomrule[1.0pt]
\end{tabular}%
}
\end{table*}

\begin{table*}[htb]
\centering
\caption{Quantitative results of different approaches on Semantic3D (\textit{semantic-8}) \cite{Semantic3D}. This test consists of 2,091,952,018 points. The scores are obtained from the recent publications. Accessed on 1 May 2020.}
\label{tab:semantic-8}
\resizebox{\textwidth}{!}{%
\begin{tabular}{rcccccccccc}
\toprule[1.0pt]
Methods & mIoU(\%) & OA(\%) & man-made. & natural. & high veg. & low veg. & buildings & hard scape & scanning art. & cars \\
\toprule[1.0pt]
TML-PC~\cite{montoya2014mind} & 39.1	& 74.5	& 80.4	& 66.1	& 42.3	& 41.2	& 64.7	& 12.4	& 0.	& 5.8 \\
TMLC-MS~\cite{hackel2016fast} & 49.4	& 85.0	& 91.1	& 69.5	& 32.8	& 21.6	& 87.6	& 25.9	& 11.3	& 55.3 \\
PointNet++~\cite{qi2017pointnet++} & 63.1	& 85.7	& 81.9	& 78.1	& 64.3	& 51.7	& 75.9	& 36.4	& 43.7	& 72.6 \\
EdgeConv~\cite{contreras2019edge} & 64.4	& 89.6	& 91.1	& 69.5	& 65.0	& 56.0	& 89.7	& 30.0	& 43..8	& 69.7 \\
SnapNet~\cite{snapnet} & 67.4	& 91.0	& 89.6	& 79.5	& 74.8	& 56.1	& 90.9	& 36.5	& 34.3	& 77.2 \\
PointGCR~\cite{ma2020global} & 69.5	& 92.1	& 93.8	& 80.0	& 64.4	& 66.4	& 93.2	& 39.2	& 34.3	& 85.3 \\
RGNet~\cite{RGNet}	& 72.0	& 90.6  & 86.4	& 70.3	& 69.5  & 68.0	& 96.9	& 43.4	& 52.3	& 89.5 \\
LCP ~\cite{boulch2020lightconvpoint} & 74.6	& 94.1	& 94.7	& 85.2	& 77.4	& 70.4	& 94.0	& 52.9	& 29.4	& 92.6 \\
SPGraph~\cite{landrieu2018large} & 76.2	& 92.9	& 91.5	& 75.6	& \textbf{78.3}	& 71.7	& 94.4	& \textbf{56.8}	& 52.9	& 88.4 \\
ConvPoint~\cite{boulch2017unstructured} & \textbf{76.5}	& 93.4	& 92.1	& 80.6	& 76.0	& \textbf{71.9}	& \textbf{95.6}	& 47.3	& \textbf{61.1}	& 87.7 \\
\textbf{\nickname{} (Ours)} & 75.8 & \textbf{95.0} & \textbf{97.4} & \textbf{93.0} & 70.2 & 65.2 & 94.4 & 49.0 & 44.7 & \textbf{92.7}
\\
\bottomrule[1.0pt]
\end{tabular}
}
\end{table*}


\begin{table*}[thb]
\centering
\caption{Quantitative results of different approaches on SemanticKITTI \cite{behley2019semantickitti}. The scores are obtained from the recent publications. Accessed on 1 May 2020.}
\label{tab:SemanticKITTI}
\resizebox{\textwidth}{!}{%
\begin{tabular}{rcccccccccccccccccccccc}
\toprule[1.0pt]
Methods & Size & \rotatebox{90}{\textbf{mIoU(\%)}} & \rotatebox{90}{Params(M)} & \rotatebox{90}{road} & \rotatebox{90}{sidewalk} & \rotatebox{90}{parking} & \rotatebox{90}{other-ground} & \rotatebox{90}{building} & \rotatebox{90}{car} & \rotatebox{90}{truck} & \rotatebox{90}{bicycle} & \rotatebox{90}{motorcycle} & \rotatebox{90}{other-vehicle} & \rotatebox{90}{vegetation} & \rotatebox{90}{trunk} & \rotatebox{90}{terrain} & \rotatebox{90}{person} & \rotatebox{90}{bicyclist} & \rotatebox{90}{motorcyclist} & \rotatebox{90}{fence} & \rotatebox{90}{pole} & \rotatebox{90}{traffic-sign} \\
\toprule[1.0pt]
PointNet \cite{qi2017pointnet} & \multirow{5}{*}{50K pts} & 14.6 & 3  & 61.6 & 35.7 & 15.8 & 1.4 & 41.4 & 46.3 & 0.1 & 1.3 & 0.3 & 0.8 & 31.0 & 4.6 & 17.6 & 0.2 & 0.2 & 0.0 & 12.9 & 2.4 & 3.7 \\
SPG \cite{landrieu2018large} &  & 17.4 & \textbf{0.25}  & 45.0 & 28.5 & 0.6 & 0.6 & 64.3 & 49.3 & 0.1 & 0.2 & 0.2 & 0.8 & 48.9 & 27.2 & 24.6 & 0.3 & 2.7 & 0.1 & 20.8 & 15.9 & 0.8 \\
SPLATNet \cite{su2018splatnet} &  & 18.4 & 0.8  & 64.6 & 39.1 & 0.4 & 0.0 & 58.3 & 58.2 & 0.0 & 0.0 & 0.0 & 0.0 & 71.1 & 9.9 & 19.3 & 0.0 & 0.0 & 0.0 & 23.1 & 5.6 & 0.0 \\
PointNet++ \cite{qi2017pointnet++} &  & 20.1 & 6 & 72.0 & 41.8 & 18.7 & 5.6 & 62.3 & 53.7 & 0.9 & 1.9 & 0.2 & 0.2 & 46.5 & 13.8 & 30.0 & 0.9 & 1.0 & 0.0 & 16.9 & 6.0 & 8.9 \\
TangentConv \cite{tangentconv} &  & 40.9 & 0.4 & 83.9 & 63.9 & 33.4 & 15.4 & 83.4 & 90.8 & 15.2 & 2.7 & 16.5 & 12.1 & 79.5 & 49.3 & 58.1 & 23.0 & 28.4 & 8.1 & 49.0 & 35.8 & 28.5 \\
\toprule[1.0pt]
SqueezeSeg \cite{wu2018squeezeseg} & \multirow{7}{*}{\begin{tabular}[c]{@{}c@{}}64*2048\\ pixels\end{tabular}} & 29.5 & 1 & 85.4 & 54.3 & 26.9 & 4.5 & 57.4 & 68.8 & 3.3 & 16.0 & 4.1 & 3.6 & 60.0 & 24.3 & 53.7 & 12.9 & 13.1 & 0.9 & 29.0 & 17.5 & 24.5 \\
SqueezeSegV2 \cite{wu2019squeezesegv2} & & 39.7 & 1  & 88.6 & 67.6 & 45.8 & 17.7 & 73.7 & 81.8 & 13.4 & 18.5 & 17.9 & 14.0 & 71.8 & 35.8 & 60.2 & 20.1 & 25.1 & 3.9 & 41.1 & 20.2 & 36.3 \\
DarkNet21Seg \cite{behley2019semantickitti} & & 47.4 & 25  & 91.4 & 74.0 & 57.0 & 26.4 & 81.9 & 85.4 & 18.6 & 26.2 & 26.5 & 15.6 & 77.6 & 48.4 & 63.6 & 31.8 & 33.6 & 4.0 & 52.3 & 36.0 & 50.0 \\
DarkNet53Seg \cite{behley2019semantickitti} & & 49.9 & 50 & \textbf{91.8} & 74.6 & 64.8 & \textbf{27.9} & 84.1 & 86.4 & 25.5 & 24.5 & 32.7 & 22.6 & 78.3 & 50.1 & 64.0 & 36.2 & 33.6 & 4.7 & 55.0 & 38.9 & 52.2 \\
RangeNet53++ \cite{rangenet++} & & 52.2 & 50  & \textbf{91.8} & \textbf{75.2} & \textbf{65.0} & 27.8 & 87.4 & 91.4 & 25.7 & 25.7 & \textbf{34.4} & 23.0 & 80.5 & 55.1 & 64.6 &38.3  & 38.8  & 4.8 & 58.6 & 47.9 & \textbf{55.9} \\
SalsaNext \cite{salsanext} & & 54.5 & 6.73 & 90.9 & 74.0 & 58.1 & 27.8 & 87.9 & 90.9 & 21.7 & 36.4 & 29.5 & 19.9 & 81.8 & 61.7 & 66.3 & \textbf{52.0} & \textbf{52.7} & 16.0 & 58.2 & 51.7 & 58.0 \\
SqueezeSegV3 \cite{xu2020squeezesegv3} & &\textbf{55.9} &26  & 91.7 & 74.8 & 63.4 & 26.4 & 89.0 & 92.5 & 29.6 & 38.7 & 36.5 & 33.0 & 82.0 & 58.7 & 65.4 & 45.6 & 46.2 & 20.1 & 59.4 & 49.6 & 58.9 \\
LatticeNet \cite{rosu2019latticenet} &- & 52.2 & - & 88.8 & 73.8 & 64.6 & 25.6 & 86.9 & 88.6 & 43.3 & 12.0 & 20.8 & 24.8 & 76.4 & 57.9 & 54.7 & 34.2 & 39.9 & \textbf{60.9} & 55.2 & 41.5 & 42.7\\
PolarNet \cite{zhang2020polarnet} &- & 54.3 & 14  & 90.8 & 74.4 &61.7 &21.7 &\textbf{90.0} &93.8 & 22.9 &40.2 & 30.1 &28.5 &\textbf{84.0} &\textbf{65.5} &67.8 &43.2 & 40.2 &5.6 &\textbf{61.3} & \textbf{51.8} & \textbf{57.5}\\
\toprule[1.0pt]
\textbf{\nickname{} (Ours)} & 50K pts & \textbf{55.9} & 1.24  & 90.5 & 74.0 & 61.8 & 24.5 & 89.7 & \textbf{94.2} & \textbf{43.9} & \textbf{47.4} & 32.2 & \textbf{39.1} & 83.8 & 63.6 & \textbf{68.6} & 48.4 & 47.4 & 9.4 & 60.4 & 51.0 & 50.7 \\
\toprule[1.0pt]
\end{tabular}%
}
\end{table*}

\begin{figure*}[thb]
\centering
\includegraphics[width=1\textwidth]{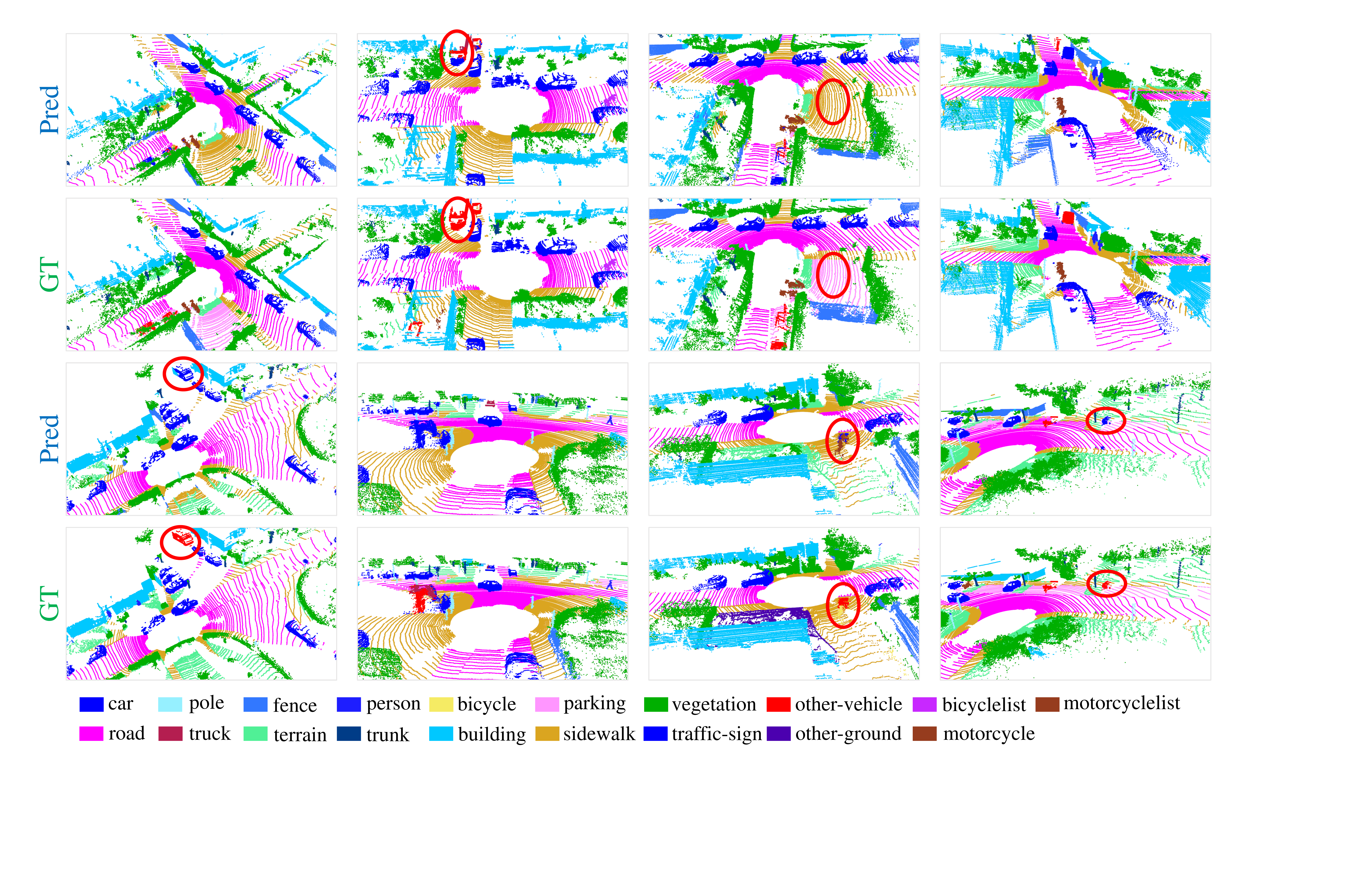}
\caption{Qualitative results of \nickname{} on the validation set of SemanticKITTI \cite{behley2019semantickitti}. Red boxes show the failure cases.}
\label{fig:semanticKITTI}
\end{figure*}

In this section, we systematically evaluate the overall efficiency of our \nickname{} on real-world large-scale point clouds for semantic segmentation. Particularly, we evaluate \nickname{} on the SemanticKITTI \cite{behley2019semantickitti} dataset, showing the detailed runtime of each component. We also evaluate the time consumption of recent representative works \cite{qi2017pointnet, qi2017pointnet++, landrieu2018large} on the same dataset. For a fair comparison, we feed the same number of points (i.e., 12$\times$81920, the batch size is 12) from the sequence into each neural network.

In addition, we also evaluate the memory consumption of \nickname{} and the baselines. In particular, we report the peak memory (GPU) of each network with the same input point clouds during inference. Note that, all experiments are conducted on the same machine with an AMD 3700X @3.6GHz CPU and an NVIDIA RTX2080Ti GPU.

\textbf{Analysis.} Table \ref{tab:efficiency} quantitatively shows the total time and memory consumption of different approaches. It can be seen that, 1) SPG \cite{landrieu2018large} has the lowest GPU memory consumption, but takes the longest time to process the point clouds due to the expensive geometrical partitioning and super-graph construction steps; 2) PointNet++ \cite{qi2017pointnet++} is also computationally expensive mainly because of the FPS sampling operation; 3) PointNet \cite{qi2017pointnet} are unable to take extremely large-scale point clouds (e.g. $10^6$ points) in a single pass due to their memory inefficient operations. 4) Thanks to the simple random sampling together with the efficient MLP-based local feature aggregator, our \nickname{} takes around 0.5s to infer the semantic labels of large-scale point clouds with nearly 1 million points (12$\times$81920 points).

\subsection{Semantic Segmentation on Benchmarks}
\label{subsec:sem_seg_benchmarks}

In this section, we evaluate the semantic segmentation of our \nickname{} on multiple large-scale public datasets: the outdoor Semantic3D \cite{Semantic3D}, SemanticKITTI \cite{behley2019semantickitti}, Toronto-3D \cite{Toronto3D}, NPM3D \cite{NPM3D}, and the indoor S3DIS \cite{2D-3D-S}. For a fair comparison, we follow KPConv \cite{thomas2019kpconv} to pre-process the whole input point clouds by using the grid-sampling strategy at the beginning. In fact, this is likely to reduce the point density and help increase the actual receptive field for each 3D point.

\textbf{(1) Evaluation on Semantic3D} \\
The Semantic3D dataset \cite{Semantic3D} consists of 15 point clouds for training and 15 for online testing. Each point cloud has up to $10^{8}$ points, covering up to 160$\times$240$\times$30 meters in real-world 3D space. The raw 3D points belong to 8 classes and contain 3D coordinates, RGB information, and intensity. We only use the 3D coordinates and color information to train and test our \nickname{}. Overall Accuracy (OA) and mean Intersection-over-Union (mIoU) of all classes are used as the evaluation metrics. For a fair comparison, we only include the results of recently published strong baselines \cite{snapnet, tchapmi2017segcloud, RF_MSSF, msdeepvoxnet, zhang2019shellnet, GACNet, landrieu2018large, thomas2019kpconv} and the current state-of-the-art approach RGNet \cite{RGNet}.

\qy{Table \ref{tab:reduced-8} presents the quantitative results of different approaches achieved on the \textit{reduced-8} test set. \nickname{} clearly outperforms all existing methods in terms of both mIoU and OA. Notably, \nickname{} also achieves superior performance on six of the eight classes, except \textit{low vegetation} and \textit{scanning artefact}. We also report the results on \textit{semantic-8} subset on Table \ref{tab:semantic-8}. \bo{Note that, compared with the \textit{reduced-8} subset, the \textit{semantic-8} subset has 10$\times$ more points and much stronger category imbalance, and therefore it is more challenging.} Our \nickname{} also achieves the best performance in terms of overall accuracy, but ranks third in mIoU performance, primary because ConvPoint \cite{conv_pts} and SPGraph \cite{landrieu2018large} achieves much better results on \textit{hard scape} and \textit{scanning artifact}. Figure \ref{fig:reduced-8-vis} shows the qualitative results of \nickname{} on this dataset.}

\textbf{(2) Evaluation on SemanticKITTI} \\
SemanticKITTI \cite{behley2019semantickitti} consists of 43552 densely annotated LIDAR scans belonging to 22 sequences. Each scan is a large-scale point cloud with $\sim 10^{5}$ points and spanning up to 160$\times$160$\times$20 meters in 3D space. Officially, the sequences 00$\sim$07 and 09$\sim$10 (19130 scans) are used for training, the sequence 08 (4071 scans) for validation, and the sequences 11$\sim$21 (20351 scans) for online testing. The raw 3D points only have 3D coordinates without color information. The mIoU score over 19 categories is used as the standard metric.

Table 3 shows a quantitative comparison of our \nickname{} with two families of recent approaches, i.e. 1) point-based methods \cite{qi2017pointnet,landrieu2018large,su2018splatnet,qi2017pointnet++,tangentconv} and 2) projection based approaches \cite{wu2018squeezeseg, wu2019squeezesegv2,behley2019semantickitti, rangenet++, rosu2019latticenet}, and Figure \ref{fig:semanticKITTI} shows some qualitative results of \nickname{} on the validation split. It can be seen that our \nickname{} surpasses all point based approaches \cite{qi2017pointnet,landrieu2018large,su2018splatnet,qi2017pointnet++,tangentconv} by large margins, with remarkable improvement of 15\% compared with the second-best approach. We also outperform most projection based methods \cite{wu2018squeezeseg,wu2019squeezesegv2,behley2019semantickitti}, but not significantly, primarily because SqueezeSegV3 \cite{xu2020squeezesegv3} achieves much better results on the small object category such as \textit{motorcyclist}. 
However, our \nickname{} is more lightweight with fewer parameters compared with the projection based methods.

\begin{table*}[thb]
\centering
\caption{Quantitative results of different approaches on Toronto3D \cite{Toronto3D}. The scores of the  baselines are obtained from \cite{Toronto3D}. Accessed on 1 May 2020.}
\label{tab:Toronto3D}
\begin{tabular*}{0.95\textwidth}{rcccccccccc}
\toprule 
Methods & OA(\%) & mIoU(\%) & Road & Rd mrk. & Natural & Building & Util. line & Pole & Car & Fence \\ \toprule 
PointNet++ \cite{qi2017pointnet++} & 84.88 & 41.81 & 89.27 & 0.00 & 69.06 & 54.16 & 43.78 & 23.30 & 52.00 & 2.95  \\
PointNet++ (MSG) \cite{qi2017pointnet++} & 92.56 & 59.47 & 92.90 & 0.00 & 86.13 & 82.15 & 60.96 & 62.81 & 76.41 & 14.43 \\
DGCNN \cite{dgcnn} & 94.24 & 61.79 & 93.88 & 0.00 & 91.25 & 80.39 & 62.40 & 62.32 & 88.26 & 15.81  \\
KPFCNN \cite{thomas2019kpconv} & 95.39 & 69.11 & 94.62 & 0.06 & 96.07 & 91.51 & 87.68 & 81.56 & 85.66 & 15.72 \\
MS-PCNN \cite{ma2019multi} & 90.03 & 65.89 & 93.84 & 3.83 & 93.46 & 82.59 & 67.80 & 71.95 & 91.12 & 22.50 \\
TGNet \cite{li2019tgnet} & 94.08 & 61.34 & 93.54 & 0.00 & 90.83 & 81.57 & 65.26 & 62.98 & 88.73 & 7.85  \\ 
MS-TGNet \cite{Toronto3D} & \textbf{95.71} & 70.50 & 94.41 & 17.19 & 95.72 & 88.83 & 76.01 & 73.97 & 94.24 & 23.64 \\ 
\textbf{\nickname{} (Ours, w/o RGB)}  & 92.95 & 77.71 & 94.61 & 42.62 & 96.89 & 93.01 & 86.51 & \textbf{78.07} & 92.85 & 37.12 \\ \toprule
\textbf{\nickname{} (Ours, w/ RGB)}$^\dagger$ & 94.37 & \textbf{81.77} & \textbf{96.69} & \textbf{64.21} & \textbf{96.92} & \textbf{94.24} & \textbf{88.06} & 77.84 & \textbf{93.37} & \textbf{42.86} \\
\bottomrule
\end{tabular*}
\end{table*}

\begin{table*}[thb]
\centering
\caption{Quantitative results of different approaches on NPM3D \cite{NPM3D}. Mean IoU (mIoU, \%), and per-class IoU (\%) are reported. The scores are obtained from the recent publications. Accessed on 1 May 2020.}
\label{tab:npmd3d}
\resizebox{0.95\textwidth}{!}{%
\begin{tabular}{rcccccccccc}
\toprule[1.0pt]
Methods & mIoU(\%) & Ground & Building & Pole & Bollard & Trash can & Barrier & Pedestrian & Car & Natural \\
\toprule[1.0pt]
RF\_MSSF~\cite{thomas2018semantic} & 56.3	& 99.3	& 88.6	& 47.8	& 67.3	& 2.3	& 27.1	& 20.6	& 74.8	& 78.8 \\
MS3\_DVS~\cite{MS3DVS}  & 66.9	& 99.0	& 94.8	& 52.4	& 38.1	& 36.0	& 49.3	& 52.6	& 91.3	& 88.6 \\
HDGCN~\cite{liang2019hierarchical} & 68.3	& 99.4	& 93.0	& 67.7	& 75.7	& 25.7	& 44.7	& 37.1	& 81.9	& 89.6 \\
MS-RRFSegNet \cite{luo2020ms}  & 79.2 & 98.6	& 98.0	& \textbf{79.7} & 74.3	& 75.1	& 57.9	& 55.9	& 82.0	& 91.4 \\
ConvPoint~\cite{boulch2017unstructured}	    & 75.9	& 99.5	& 95.1	& 71.6	& 88.7	& 46.7	& 52.9	& 53.5	& 89.4	& 85.4 \\
KPConv~\cite{thomas2019kpconv}  & 82.0	& 99.5	& 94.0	& 71.3	& 83.1	& \textbf{78.7}	& 47.7	& \textbf{78.2}	& 94.4	& 91.4 \\
LCP~\cite{boulch2020lightconvpoint}  & \textbf{82.7}	& \textbf{99.6}	& \textbf{98.1}	& \textbf{77.2}	& \textbf{91.1}	& 64.7	& \textbf{66.5}	& 58.1	& \textbf{95.6}	& \textbf{93.9} \\
\textbf{\nickname{} (Ours)} & 78.5 & 99.5 & 97.0 & 71.0 & 86.7 & 50.5 & 65.5 & 49.1 & 95.3 & 91.7 \\
\bottomrule[1.0pt]
\end{tabular}%
}
\end{table*}

\begin{table*}[thb]
\centering
\caption{Quantitative results of different approaches on S3DIS \cite{2D-3D-S} (6-fold cross-validation). Overall Accuracy (OA, \%), mean class Accuracy (mAcc, \%), mean IoU (mIoU, \%), and per-class IoU (\%) are reported.}
\label{tab:s3dis}
\resizebox{\textwidth}{!}{%
\begin{tabular}{rcccccccccccccccc}
\toprule[1.0pt]
Methods & OA(\%) & mAcc(\%)& mIoU(\%) & ceil. & floor & wall & beam & col. & wind. & door & table & chair & sofa & book. & board & clut. \\ 
\toprule[1.0pt]
PointNet \cite{qi2017pointnet} & 78.6 & 66.2 & 47.6 & 88.0 & 88.7 & 69.3 & 42.4 & 23.1 & 47.5 & 51.6 & 54.1 & 42.0 & 9.6 & 38.2 & 29.4 & 35.2 \\
RSNet \cite{RSNet} & - & 66.5 & 56.5 & 92.5 & 92.8 & 78.6 & 32.8 & 34.4 & 51.6 & 68.1 & 59.7 & 60.1 & 16.4 & 50.2 & 44.9 & 52.0 \\
3P-RNN \cite{3PRNN} & 86.9 & - & 56.3 & 92.9 & 93.8 & 73.1 & 42.5 & 25.9 & 47.6 & 59.2 & 60.4 & 66.7 & 24.8 & 57.0 & 36.7 & 51.6 \\
SPG \cite{landrieu2018large}& 86.4 & 73.0 & 62.1 & 89.9 & 95.1 & 76.4 & 62.8 & 47.1 & 55.3 & 68.4 & \textbf{73.5} & 69.2 & 63.2 & 45.9 & 8.7 & 52.9 \\
PointCNN \cite{li2018pointcnn} & 88.1 & 75.6 & 65.4 & 94.8 & 97.3 & 75.8 & 63.3 & 51.7 & 58.4 & 57.2 & 71.6 & 69.1 & 39.1 & 61.2 & 52.2 & 58.6 \\ 
PointWeb \cite{pointweb} & 87.3 & 76.2 & 66.7 & 93.5 & 94.2 & 80.8 & 52.4 & 41.3 & 64.9 & 68.1 & 71.4 & 67.1 & 50.3 & 62.7 & 62.2 & 58.5  \\
ShellNet \cite{zhang2019shellnet} & 87.1  & - & 66.8 & 90.2 & 93.6 & 79.9 & 60.4 & 44.1 & 64.9 & 52.9 & 71.6 & \textbf{84.7} & 53.8 & 64.6 & 48.6 & 59.4 \\
PointASNL \cite{yan2020pointasnl} & \textbf{88.8}  & 79.0 & 68.7 & \textbf{95.3} & \textbf{97.9} & 81.9 & 47.0 & 48.0 & \textbf{67.3} & 70.5 & 71.3 & 77.8 & 50.7 & 60.4 & 63.0 & \textbf{62.8} \\
KPConv\_ \textit{rigid} \cite{thomas2019kpconv} & - & 78.1 & 69.6 & 93.7 & 92.0 & 82.5 & 62.5 & 49.5  & 65.7 & \textbf{77.3} & 57.8 & 64.0 & 68.8 & 71.7 & 60.1 & 59.6 \\
KPConv\_ \textit{deform} \cite{thomas2019kpconv} &- & 79.1 &\textbf{70.6} &93.6 &92.4 & \textbf{83.1} & \textbf{63.9} & \textbf{54.3} & 66.1 & 76.6 &57.8 &64.0 & \textbf{69.3} & \textbf{74.9} &61.3 & 60.3 \\
\textbf{\nickname{} (Ours)} & 88.0 & \textbf{82.0} & 70.0 & 93.1 & 96.1 & 80.6 & 62.4 & 48.0 & 64.4 & 69.4 & 69.4 & 76.4 & 60.0 & 64.2 & \textbf{65.9} & 60.1 \\
\bottomrule[1.0pt]
\end{tabular}%
}
\end{table*}

\textbf{(3) Evaluation on Toronto-3D} \\
The Toronto-3D dataset \cite{Toronto3D} consists of 78.3 million points acquired by the mobile laser systems (MLS), which covers approximately 1KM of urban outdoor 3D space. Following \cite{Toronto3D}, the section \textit{L002} is used for testing and the remaining three sections are used for training. OA and mIoU over 8 categories are used as evaluation metrics. 

\begin{figure}[thb]
\centering
\includegraphics[width=0.47\textwidth]{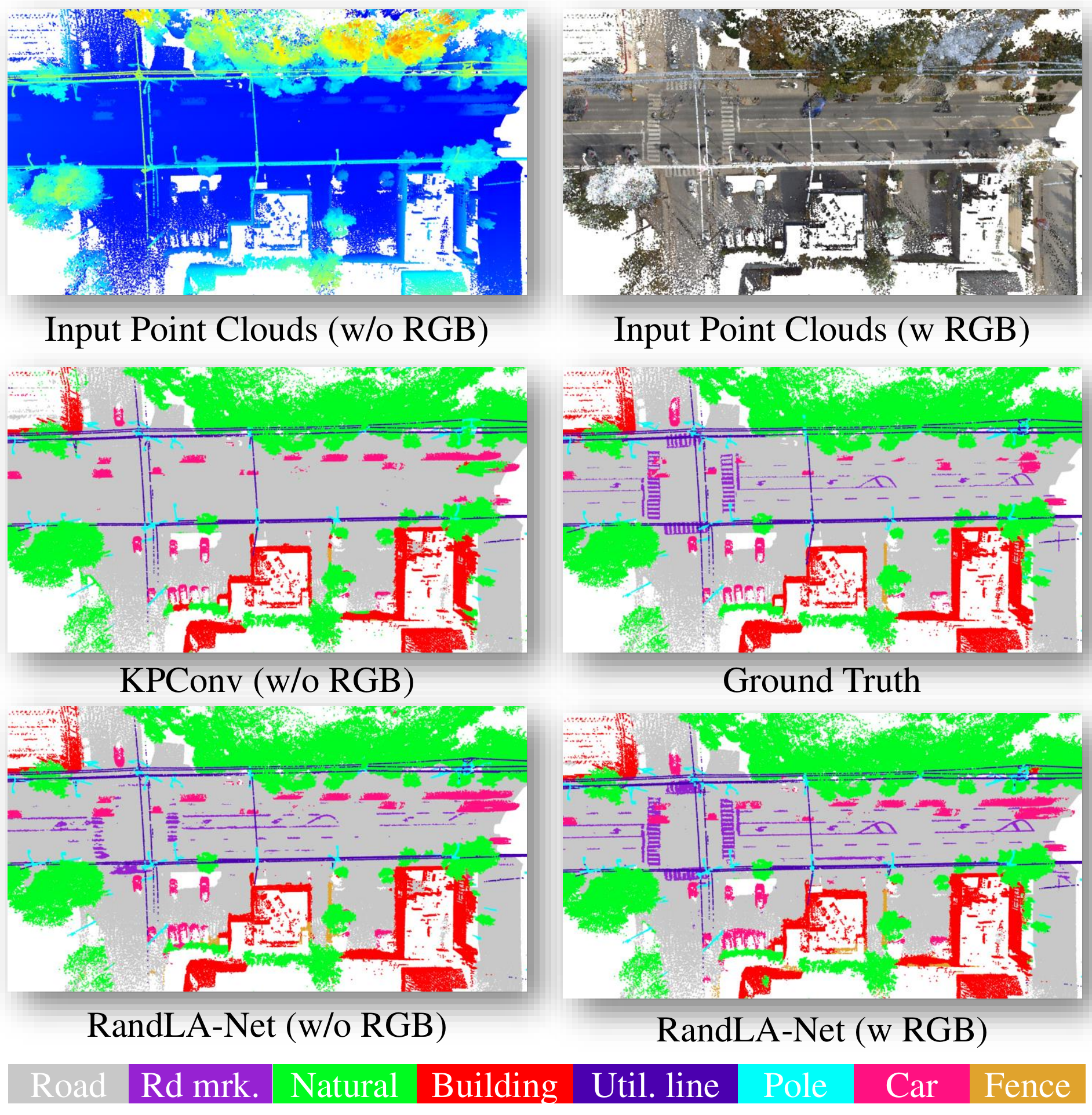}
\caption{Semantic segmentation results of KPConv \cite{thomas2019kpconv} and our approach on Toronto-3D \cite{Toronto3D}. Our method clearly performs better on the category of road marking.
\label{fig:Toronto-3d}}
\end{figure}

Table \ref{tab:Toronto3D} reports the quantitative results achieved by our \nickname{} with several recently published baselines \cite{qi2017pointnet++, dgcnn, thomas2019kpconv, ma2019multi, li2019tgnet, Toronto3D}, and Figure \ref{fig:Toronto-3d} shows the qualitative results. Note that, only the 3D coordinates of each point are feed into the network for fair comparison \cite{Toronto3D}. It can be seen that our \nickname{} achieves the best results on this dataset in terms of mIoU, improving the previous state-of-the-art MS-TGNet \cite{Toronto3D} from 70.50\% to 77.71\%. 

\bo{It is observed that \textit{Road marking} is the hardest to be recognized by all approaches. Primarily, this is because the pure geometric patterns of \textit{Road marking} can not be easily distinguished against its neighbourhood. Particularly, given only 3D coordinates of points, it is virtually impossible to differentiate the \textit{Road marking} and \textit{Road}, as illustrated in Figure \ref{fig:Toronto-3d}. This motivates us to include the appearance of points to further improve the accuracy. We simply train our \nickname{} with both 3D coordinates and color information as input. As shown in the last row of Table \ref{tab:Toronto3D}, the IoU score of \textit{Road marking} has been significantly improved by 22\%, demonstrating the importance of appearance in this difficult case.
}

\qy{\textbf{(4) Evaluation on NPM3D} \\
The NPM3D dataset \cite{NPM3D} is a large-scale and high-quality urban point clouds dataset. Similar to the Toronto-3D dataset, the NPM3D dataset also consists of around 2KM MLS point cloud acquired in two cities. The mean IoU (mIoU) of the total 10 classes are used as the main evaluation metric.}

\qy{Table \ref{tab:npmd3d} quantitatively shows the results achieved by our \nickname{} and exiting baselines \cite{conv_pts, boulch2020lightconvpoint, thomas2019kpconv, MS3DVS, RF_MSSF,liang2019hierarchical}. In this dataset, only the 3D coordinates of each point is available. 
\bo{Due to the relatively lower scores of two minority classes \textit{Trash can} and \textit{Pedestrian}}, our \nickname{} achieves a mean IoU score of 78.5\%, ranking after the latest LightConvPoint \cite{boulch2020lightconvpoint} and KPConv \cite{thomas2019kpconv}. 
\bo{Primarily, the point features of \textit{Trash can} and \textit{Pedestrian} are very likely to be randomly dropped in our \nickname{} 
because the two categories only have extremely small number of points scanned.
}
}

\textbf{(5) Evaluation on S3DIS} \\
The S3DIS dataset \cite{2D-3D-S} consists of 271 rooms belonging to 6 large areas. Each point cloud is a medium-sized single room ($\sim$ 20$\times$15$\times$5 meters) with dense 3D points. To evaluate the semantic segmentation of our \nickname{}, we use the standard 6-fold cross-validation in our experiments. The mean IoU (mIoU), mean class Accuracy (mAcc) and Overall Accuracy (OA) of the total 13 classes are compared.

\bo{Table \ref{tab:s3dis} quantitatively compares the performance of our \nickname{} with existing baselines on this dataset}. Our \nickname{} achieves on-par or better performance than state-of-the-art methods. Note that, most of these baselines \cite{qi2017pointnet++, li2018pointcnn, pointweb, zhang2019shellnet, dgcnn, chen2019lsanet} tend to use sophisticated but expensive operations or samplings to optimize the networks on small blocks (e.g., 1$\times$1 meter) of point clouds, and the relatively small rooms act in their favours to be divided into tiny blocks. By contrast, \nickname{} is able to take the entire rooms as input and efficiently infer per-point semantics in a single pass.

\subsection{Ablation Studies}
\label{sec:ablation}
\begin{table}[thb]
\centering
\caption{The mean IoU scores of all ablated networks based on our full \nickname{}.}
\label{tab:ablative}
\resizebox{0.4\textwidth}{!}{%
\begin{tabular}{lc}
\Xhline{2\arrayrulewidth}
Ablated networks & mIoU(\%) \\
\Xhline{2\arrayrulewidth}
(1) Remove local spatial encoding & 49.8 \\
(2) Replace with max-pooling & 55.2 \\
(3) Replace with mean-pooling & 53.4 \\
(4) Replace with sum-pooling & 54.3 \\
(5) Simplify dilated residual block & 48.8 \\
\textbf{(6) The Full framework (\nickname{})} & \textbf{57.1} \\
\Xhline{2\arrayrulewidth}
\end{tabular}%
}
\vspace{-0.2cm}
\end{table}

\begin{table}[thb]
\centering
\caption{The mean IoU scores of our \nickname{} with different designs of LocSE.}
\label{tab:LocSE_ablation}
\resizebox{0.46\textwidth}{!}{%
\begin{tabular}{lc}
\Xhline{2\arrayrulewidth}
LocSE & mIoU(\%) \\
\Xhline{2\arrayrulewidth}
(1) $(p_{i})$ &  48.9\\
(2) $(p_{i}^{k})$ &  50.7\\
(3) $(p_{i}-p_{i}^{k})$ &  56.4\\
(4) $(||p_i-p_i^k||)$ &  35.9\\
(5) $(p_{i}, p_{i}^{k})$ & 52.5 \\
(6) $(p_{i}, p_{i}^{k}, p_{i}-p_{i}^{k})$ &  56.8\\
(7) $(p_{i}, p_{i}^{k}, ||p_i-p_i^k||)$ & 53.7 \\
(8)  $(p_{i}-p_{i}^{k}, ||p_i-p_i^k||)$ &  56.9 \\
\textbf{(9) $(p_{i}, p_{i}^{k}, p_{i}-p_{i}^{k}, ||p_i-p_i^k||)$ (The Full Unit)} & \textbf{57.1} \\ 
\Xhline{2\arrayrulewidth}
\end{tabular}
}
\end{table}

Since the impact of random sampling is fully studied in Section \ref{sec:eff_sampling}, we  conduct the following ablation studies for our local feature aggregation module. All ablated networks are trained on sequences 00$\sim$07 and 09$\sim$10, and tested on the sequence 08 of SemanticKITTI dataset \cite{behley2019semantickitti}.

\subsubsection{Ablation of \nickname{} Framework}\label{ablation:framework}
Since our \nickname{} stacks multiple of our key components: LocSE, Attentive Pooling and Dilated Residual Block, we therefore conduct the following 5 ablation studies to demonstrate the effectiveness of each component.

\textit{(1) Removing local spatial encoding (LocSE).} This unit enables each 3D point to explicitly observe its local geometry. After removing locSE, we directly feed the local point features into the subsequent attentive pooling.

\textit{(2$\sim$4) Replacing attentive pooling by max/mean/sum pooling.} The attentive pooling unit learns to automatically combine all local point features. By comparison, the widely used max/mean/sum poolings tend to hard select or combine features, therefore their performance may be sub-optimal.

\textit{(5) Simplifying the dilated residual block.} The dilated residual block stacks multiple LocSE units and attentive poolings, substantially dilating the receptive field for each 3D point. By simplifying this block, we use only one LocSE unit and attentive pooling per layer, i.e. we do not chain multiple blocks as in our original \nickname{}.

Table \ref{tab:ablative} compares the mIoU scores of all ablated networks. From this, we can see that: 1) The greatest impact is caused by the removal of the chained spatial embedding and attentive pooling blocks. This is highlighted in Figure \ref{fig:Residual}, which shows how using two chained blocks allows information to be propagated from a wider neighbourhood, i.e. approximately $K^2$ points as opposed to  just $K$. This is especially critical with random sampling, which is not guaranteed to preserve a particular set of points. 2) The removal of the local spatial encoding unit shows the next greatest impact on performance, demonstrating that this module is necessary to effectively learn local and relative geometry context. 3) Removing the attention module diminishes performance by not being able to effectively retain useful features. From this ablation study, we can see how the proposed neural units complement each other to attain our state-of-the-art performance.

\subsubsection{Ablation of LocSE}
As designed in Section \ref{Attentive_aggregation}, our LocSE component encodes the relative point position based on the following equation:
\begin{equation}
  \mathbf{r}_{i}^{k} = MLP\Big(p_i \oplus p_i^k \oplus (p_i-p_i^k) \oplus ||p_i-p_i^k||\Big)
\label{Eq1_sup}
\end{equation}

We further investigate the effects of different spatial information in our framework. Particularly, we conduct the following 8 ablative experiments for LocSE:

\textit{(1-4) Encoding the point $p_i$ only, the neighboring points $p_i^k$ only, the relative position $p_{i}-p_{i}^{k}$ only, and the Euclidean distance $||p_i-p_i^k||$ only, respectively.}

\textit{(5) Encoding the point $p_i$ and its neighboring points $p_i^k$}. 

\textit{(6) Encoding the point $p_i$, neighboring points $p_i^k$, and their relative position $p_{i}-p_{i}^{k}$}.

\textit{(7) Encoding the point $p_i$, neighboring points $p_i^k$, and their Euclidean distance $||p_i-p_i^k||$}.

\textit{(8) Encoding only the relative position $p_{i}-p_{i}^{k}$ and their Euclidean distance $||p_i-p_i^k||$.}

Table \ref{tab:LocSE_ablation} compares the mIoU scores of our network with different designs of LocSE. We can see that: 1) Explicitly encoding all spatial information leads to the highest mIoU scores. 2) The relative position $p_{i}-p_{i}^{k}$ plays an important role in the LocSE component, primarily because the relative point position enables the network to be aware of the local geometric patterns. 3) Only encoding the point position $p_i$, $p_i^k$ or Euclidean distance $||p_i-p_i^k||$ is sub-optimal, because the individual point positions and distance only are less informative for capturing the local geometric patterns.

\subsubsection{Ablation of Dilated Residual Block}
As shown in Figure \ref{fig:network}, we stack two sets of LocSE and Attentive Pooling units as the standard dilated residual block to gradually increase the receptive field. To further evaluate how the number of aggregation units in our dilated residual block impact the entire network, we conduct the following two more ablative experiments.

\textit{(1) Using only one set of LocSE and attentive pooling}. In this setting, the receptive field for each 3D point becomes smaller. 

\textit{(2) Using three sets of LocSE and attentive pooling.} The receptive field becomes exponentially larger.

\begin{table}[htb]
\centering
\caption{The mIoU scores of \nickname{} using different number of aggregation units in each residual block.}
\label{tab:num_aggregation}
\resizebox{0.46\textwidth}{!}{%
\begin{tabular}{lc}
\Xhline{2\arrayrulewidth}
Dilated residual block & mIoU(\%) \\
\Xhline{2\arrayrulewidth}
(1) one set of LocSE and attentive pooling &  52.9 \\ 
(2) three sets of LocSE and attentive pooling &  54.2 \\ 
\textbf{(3) two sets (The Standard Block)} & \textbf{57.1} \\ 
\Xhline{2\arrayrulewidth}
\end{tabular}%
}
\end{table}

Table \ref{tab:num_aggregation} shows the mIoU scores of our \nickname{} with different number of aggregation units in each dilated residual block. It can be seen that: 1) Only one set of LocSE and attentive pooling in the dilated residual block leads to a significant drop of the mIoU score, due to the limited receptive field. 2) Three sets of LocSE and attentive pooling do not improve the accuracy as expected.
This is because the significantly increased receptive fields and the large number of trainable parameters tend to be overfitted.

\subsubsection{Ablation of Sampling Approaches}
We conduct additional experiments to evaluate the segmentation performance of our framework when adopting different sampling strategies as discussed in Section \ref{Sub-sampling}. The experimental settings are the same as in Section \ref{ablation:framework}. As shown in Table  \ref{tab:RandLA-with-sampling}, all results are achieved on the validation set (\ie   sequence 08) of the SemanticKITTI dataset and the S3DIS dataset (\textit{Area-5}), respectively.

\begin{table}[htb]
\centering
\caption{Quantitative results of our framework with different sampling strategies on the validation set of the SemanticKITTI dataset \revminor{and the S3DIS dataset (Area-5)}. The symbol - means the results are unavailable. For Continuous Relaxation based Sampling (CRS), we only adopt it at the last sampling layer in our framework due to the excessive GPU memory consumption. $^\dagger$The model did not converge.}
\label{tab:RandLA-with-sampling}
\resizebox{0.36\textwidth}{!}{%
\begin{tabular}{lcc}
\Xhline{2\arrayrulewidth}
\multirow{2}{*}{} & \multicolumn{2}{c}{mIoU (\%)} \\
 & SemanticKITTI & S3DIS \\
\Xhline{2\arrayrulewidth}
RandLA-Net+FPS & 56.5 & \textbf{64.3} \\
RandLA-Net+IDIS & 46.8 & 54.6 \\
RandLA-Net+PDS & 54.0 & 58.9 \\
RandLA-Net+GS$^\dagger$ & - & - \\
RandLA-Net+CRS & 49.4 & 56.7 \\
RandLA-Net+PGS & 55.7 & 60.3 \\
RandLA-Net+RS & \textbf{57.1} & 63.4 \\ 
\Xhline{2\arrayrulewidth}
\end{tabular}%
}
\end{table}

It can be seen that the segmentation performance of our framework is comparable when used with random sampling, farthest point sampling, or policy gradient-based sampling, showing that our local feature aggregation module is indeed effective and amenable to other sampling methods. Note that, the performance significantly drops when the inverse density importance sampling is used, primarily because the selected points have lower density and tend to be outliers and noisy in practice.

\subsection{Analysing Point Neighbours}
In our RandLA-Net, we use the simple KNN to find a fixed number of neighboring points to extract local patterns for each 3D point. Intuitively, the choice of $K$ in KNN may affect the overall performance. In addition, the spatial range of neighboring points searched by KNN varies if the point density changes, making it unable to maintain a geometrically consistent neighborhood for each point \cite{thomas2018semantic}. Alternatively, the neighboring points can be queried within a fixed spherical radius, guaranteeing the consistency of local boundaries. In this section, we dive deep into point neighbor search, evaluating the choice of $K$ and comparing KNN with radius query. 

\subsubsection{Choice of K in KNN}
\bo{Intuitively, the larger the $K$, the more complex the local geometric patterns that can be learned and the heavier computation required. We conduct 7 groups of experiments where $K$ varies from 4 to 64. The evaluation setting and protocol are the same as used in Section \ref{sec:ablation}.
}

\bo{ Figure \ref{fig:KNN} (a) shows the variation of segmentation performance given different choices of $K$ in KNN. It can be seen that, the network achieves a significant gain (mIoU score: 46\% $\rightarrow$ 57\%) when the number of neighbouring points $K$ increases from 4 to 16, after which the mean IoU score drops gradually given a larger size of $K$. This demonstrates that, 1) a very small set of neighbouring points is not helpful for the network to learn useful point local features due to the limited receptive field; 2) an extremely large set of neighbouring points is also unlikely to improve the accuracy, since the learned complex local features tend to overfit the training examples and are not general.
}

\subsubsection{KNN vs. Spherical Neighbour}
To further investigate how the neighbouring query mechanism impacts the performance, we replace the KNN with spherical radius query in our \nickname{} without modifying any other components. Similarly, we also conduct 7 groups of experiments by gradually increasing the radius of the first encoding layer from 0.1 to 0.7, and then doubling the radius after each sampling layer. For a fair comparison \revminor{and better parallelization}, we consistently select $K=16$ neighborhood points in each spherical neighborhood. \revminor{This is achieved by following PointNet++} \cite{qi2017pointnet++} \revminor{to downsample or pad within each neighboring point set.}
\begin{figure}[thb]
\centering
\includegraphics[width=0.5\textwidth]{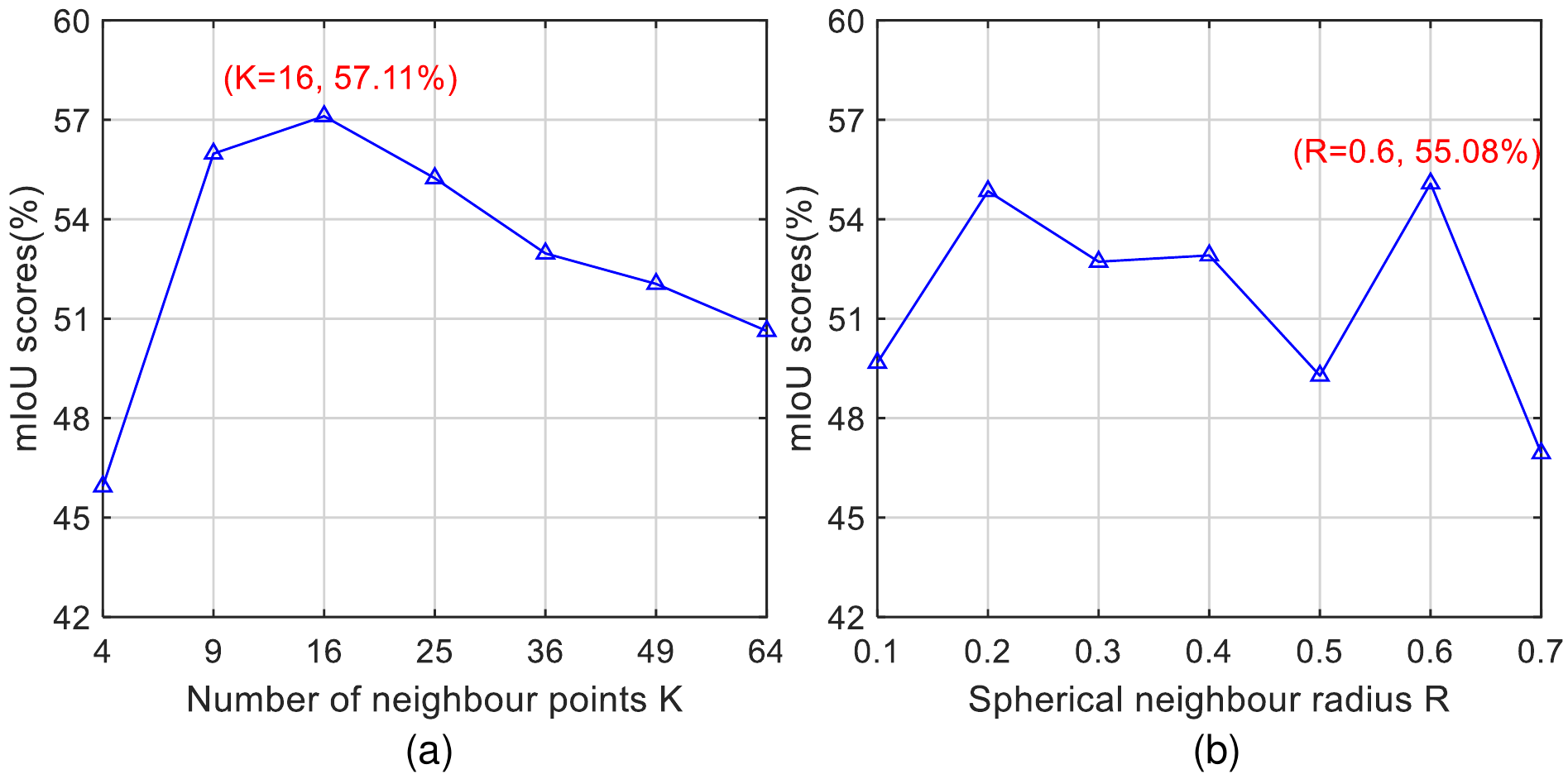}
\caption{The results of our RandLA-Net with different neighbour searching methods. Left: The mIoU scores for different choices of $K$ in KNN. Right: The mIoU scores for different choices of the radius $R$ in spherical neighbours.}
\label{fig:KNN}
\vspace{-0.2cm}
\end{figure}

As shown in Fig \ref{fig:KNN} (b), the network achieves the best performance when the  radius of the first layer is set to 0.6m, with a mean IoU score of 55.08\%. This is slightly lower than the best model with KNN (55.08\% vs 57.11\%), demonstrating that our framework is able to adapt to the spatial inconsistency potentially caused by KNN. Primarily, this is because the attentive pooling mechanism of our LocSE unit tends to automatically learn different weighs for all neighbouring points, and the spatial locations of all points have been implicitly considered.

\subsubsection{Attention Scores of Neighbouring Points}

\begin{figure}[thb]
\centering
\includegraphics[width=0.5\textwidth]{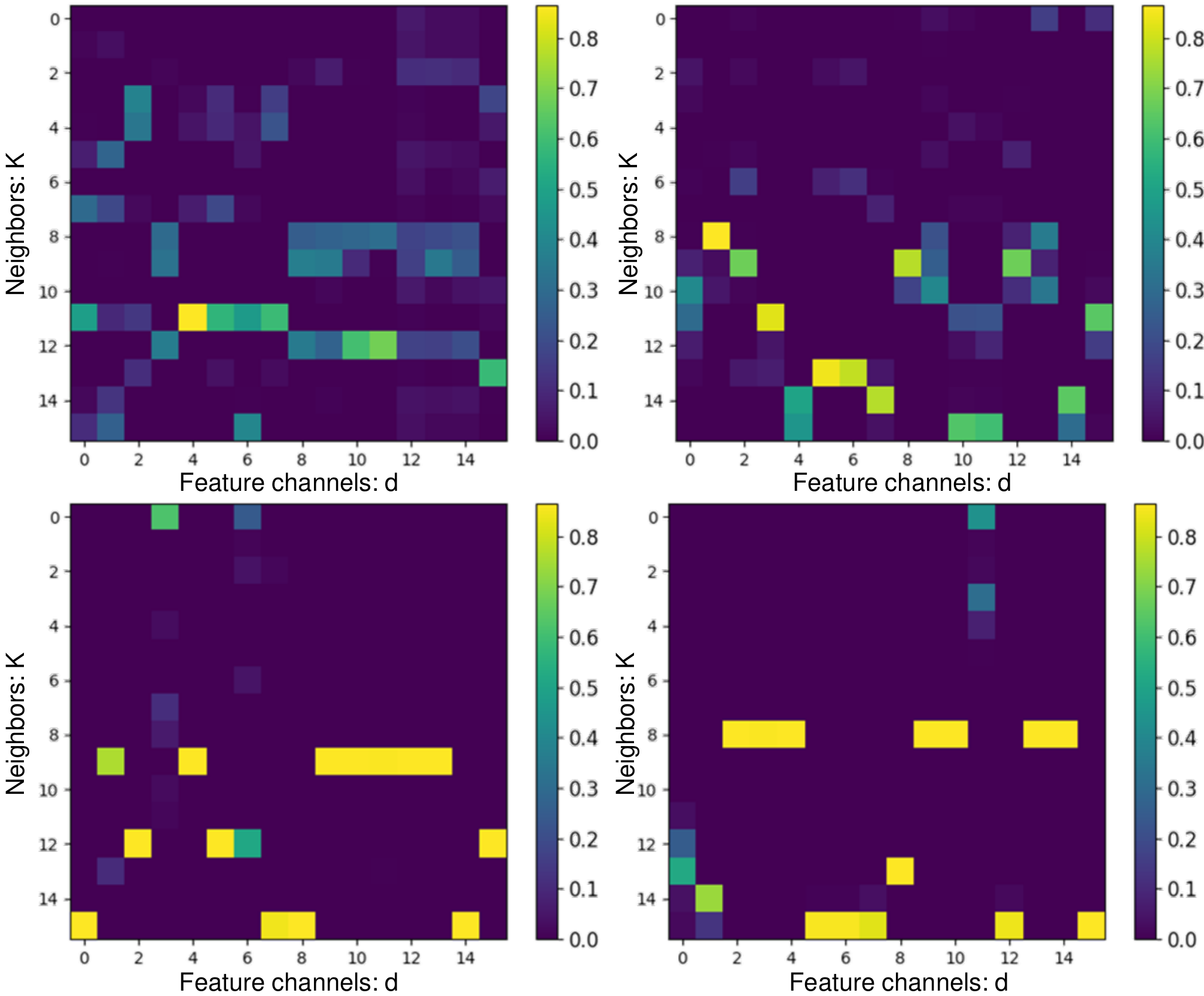}
\caption{Visualization of the learned attention scores in different encoding layers. From top left to bottom right: the learned attention score map in the first (16$\times$16), second (16$\times$64), third (16$\times$128), and the last encoding layers (16$\times$256). For better visualization, we only keep the left part of the attention score map as an 16$\times$16 image. The yellow color represents higher attention scores.}
\label{attention_matrix}
\vspace*{-0.4cm}
\end{figure}

After the neighbouring points are selected by KNN, their features are aggregated by the attentive pooling as shown in Figure \ref{fig:network}. To investigate how the features are effectively integrated, we visualize and analyze the learned attention scores in this section. As shown in Figure \ref{attention_matrix}, for a specific center point, the learned attention scores for its 16 neighboring points tend to be similar in the first encoding layer, whereas the learned attention scores for its new 16 neighboring points in the subsequent layers are dramatically different. In effect, this is because the neighbouring points at the first layer are likely to be similar and equally informative, thus their learned attention scores are similar as well. After the large-scale point cloud is significantly downsampled and progressively aggregated, the retained point features tend to be highly compact and have different semantic context. Therefore, the attention scores of those neighbouring points which do not have similar semantic meanings with the center point, tend to be zero, while the most similar neighbouring point tends to have a high attention score. 

\section{Discussion}\label{sec:discussion}
Our RandLA-Net achieves excellent performance in semantic segmentation of large-scale 3D point clouds, it also has limitations. First, the KNN used for searching point neighbours does not scale up linearly. This would be a major computation bottleneck to process extremely large-scale point clouds (e.g., $>10^6$ points). One possible solution is to leverage the point-voxel representation \cite{Point_voxel_cnn, xu2020grid} by searching neighbours within each voxel instead of the entire point cloud. Alternatively, the random sampling together with KNN can be replaced by K-random path selection from KD-Tree or OcTree. In this case, the KNN could be efficiently retrieved during sampling. Second, the random sampling used in our framework focuses on the observed points and cannot produce potentially interesting information at some missing/occluded regions. Therefore, it would be interesting to generate new random point sets to learn better local patterns. \revminor{Finally, combining the efficiency of random sampling and the effectiveness of advanced point convolutions} \cite{li2018pointcnn, thomas2019kpconv, thomas2019kpconv} \revminor{is also worth to be further investigated.}

\section{Conclusion} \label{sec:conclusion}
In this paper, we demonstrated that it is possible to efficiently and effectively segment large-scale point clouds by using a lightweight network architecture. In contrast to most current approaches, that rely on expensive sampling strategies, we instead use random sampling in our framework to significantly reduce the memory footprint and computational cost. A local feature aggregation module is also introduced to effectively preserve useful features from a wide neighbourhood. Extensive experiments on multiple benchmarks demonstrate the high efficiency and the state-of-the-art performance of our approach. It would be interesting to extend our framework for the end-to-end 3D instance segmentation on large-scale point clouds by drawing on the recent work \cite{3dbonet} and also for the real-time dynamic point cloud processing \cite{liu2019meteornet}.

\vspace*{-0.2cm}
\ifCLASSOPTIONcompsoc
  \section*{Acknowledgments}
\else
  \section*{Acknowledgment}
\fi

This work was supported in part by Amazon Web Services in the Oxford-Singapore Human-Machine Collaboration Programme. The work of Qingyong Hu was supported by the China Scholarship Council (CSC) Scholarship. The work of Yulan Guo was supported by National Natural Science Foundation of China (No. U20A20185, 61972435), the Natural Science Foundation of Guangdong Province (2019A1515011271), the Science and Technology Innovation Committee of Shenzhen Municipality (RCYX20200714114641140, JCYJ20190807152209394).

\clearpage
\bibliographystyle{IEEEtran}
\bibliography{egbib}
\ifCLASSOPTIONcaptionsoff
  \newpage
\fi

\vspace{-1cm}
\begin{IEEEbiography}[{\includegraphics[width=1in,height=1in,clip,keepaspectratio]{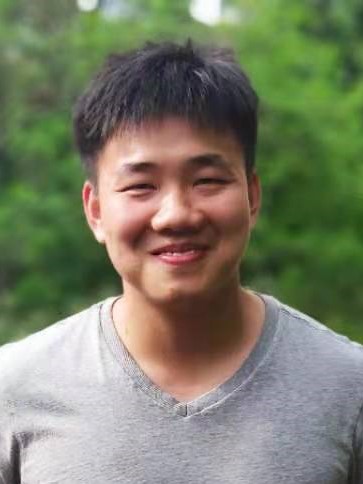}}]{Qingyong Hu} received his M.Eng. degree in information and communication engineering from the National University of Defense Technology (NUDT) in 2018. He is currently a DPhil candidate in the Department of Computer Science at the University of Oxford. His research interests lie in 3D computer vision, large-scale point cloud processing, and visual tracking.
\end{IEEEbiography}
\vspace{-1cm}
\begin{IEEEbiography}[{\includegraphics[width=1in,height=1in,clip,keepaspectratio]{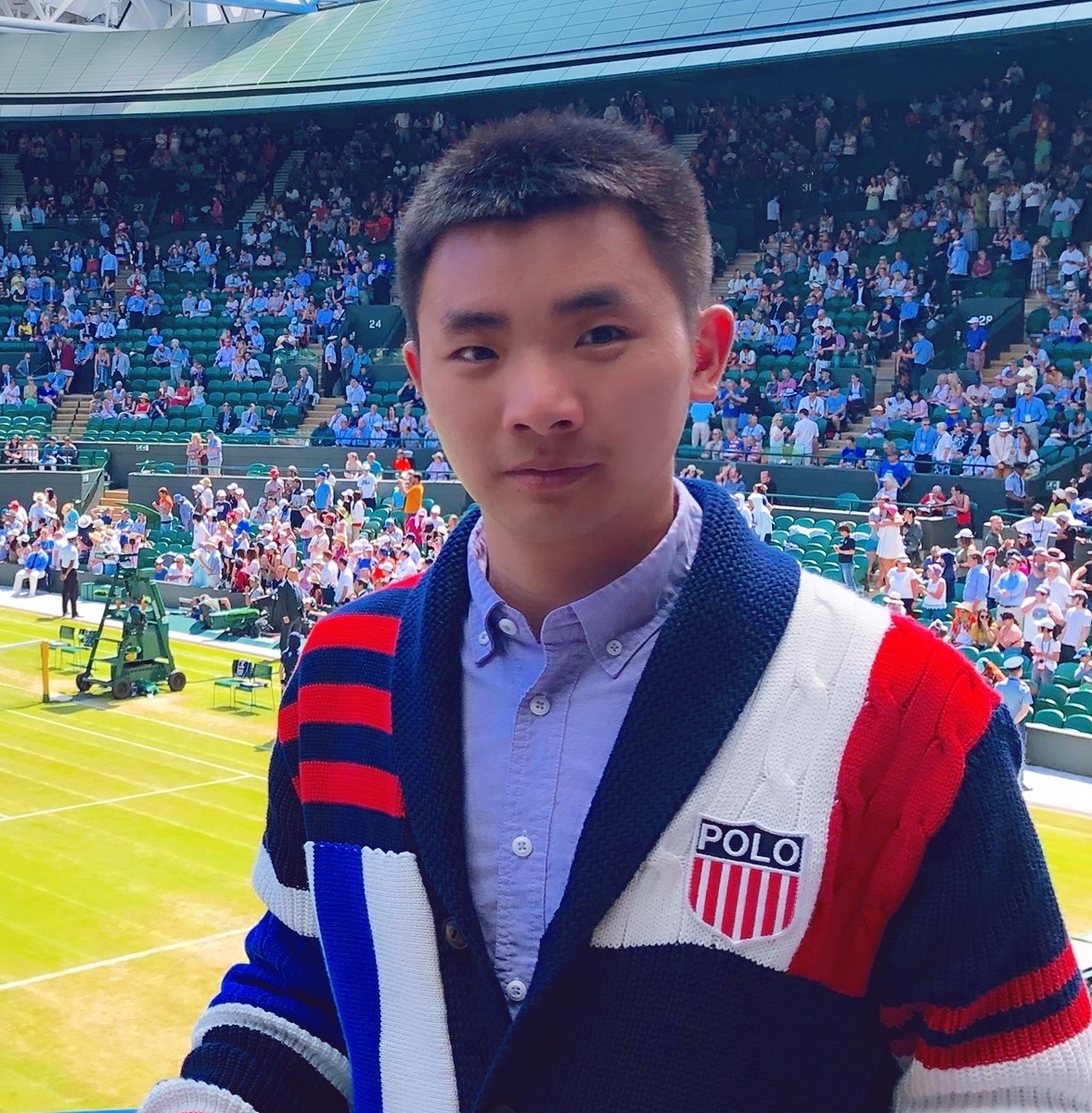}}]{Bo Yang} is an Assistant Professor in the Department of Computing at The Hong Kong Polytechnic University. He obtained his DPhil degree (2020) from the University of Oxford. His research interests lie in deep learning, computer vision and robotics.
\end{IEEEbiography}
\vspace{-1cm}
\begin{IEEEbiography}[{\includegraphics[width=1in,height=1in,clip,keepaspectratio]{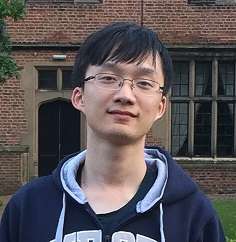}}]{Linhai Xie} is a DPhil student at the Department of Computer Science, University of Oxford. Before that, he obtained his BEng Degree at National University of Defense Technology, China. His research is focused on learning based robot perception and autonomy, including robot navigation, reinforcement learning, robotic vision and SLAM.
\end{IEEEbiography}
\vspace{-1cm}

\begin{IEEEbiography}[{\includegraphics[width=1in,height=1.25in,clip,keepaspectratio]{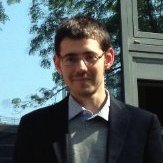}}]{Stefano Rosa} received his Ph.D. in Mechatronics Engineering from Politecnico di Torino, Italy, in 2014. He is currently a research fellow in the Department of Computer Science at University of Oxford, UK, working on long-term navigation, Human-Robot Interaction and intuitive physics.
\end{IEEEbiography}
\vspace{-1cm}

\begin{IEEEbiography}[{\includegraphics[width=1in,height=1.25in,clip,keepaspectratio]{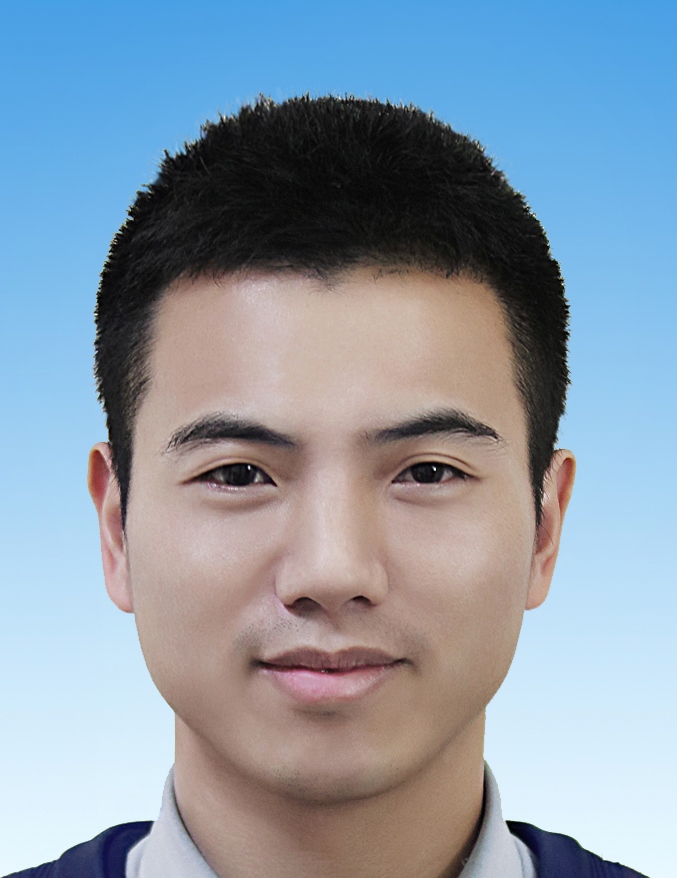}}]{Yulan Guo} is currently an associate professor. He received the B.Eng. and Ph.D. degrees from National University of Defense Technology (NUDT) in 2008 and 2015, respectively. He has authored over 100 articles in journals and conferences, such as the IEEE TPAMI and IJCV. His current research interests focus on 3D vision, particularly on 3D feature learning, 3D modeling, 3D object recognition, and scene understanding. Dr. Guo received the ACM China SIGAI Rising Star Award in 2019, the Wu-Wenjun Outstanding AI Youth Award in 2019. He served as an associate editor for IET Computer Vision and IET Image Processing, a guest editor for IEEE TPAMI, an area chair for several conferences (e.g., CVPR 2021, ICCV 2021, and ACM MM 2021).
\end{IEEEbiography}
\vspace{-1cm}

\begin{IEEEbiography}[{\includegraphics[width=1in,height=1in,clip,keepaspectratio]{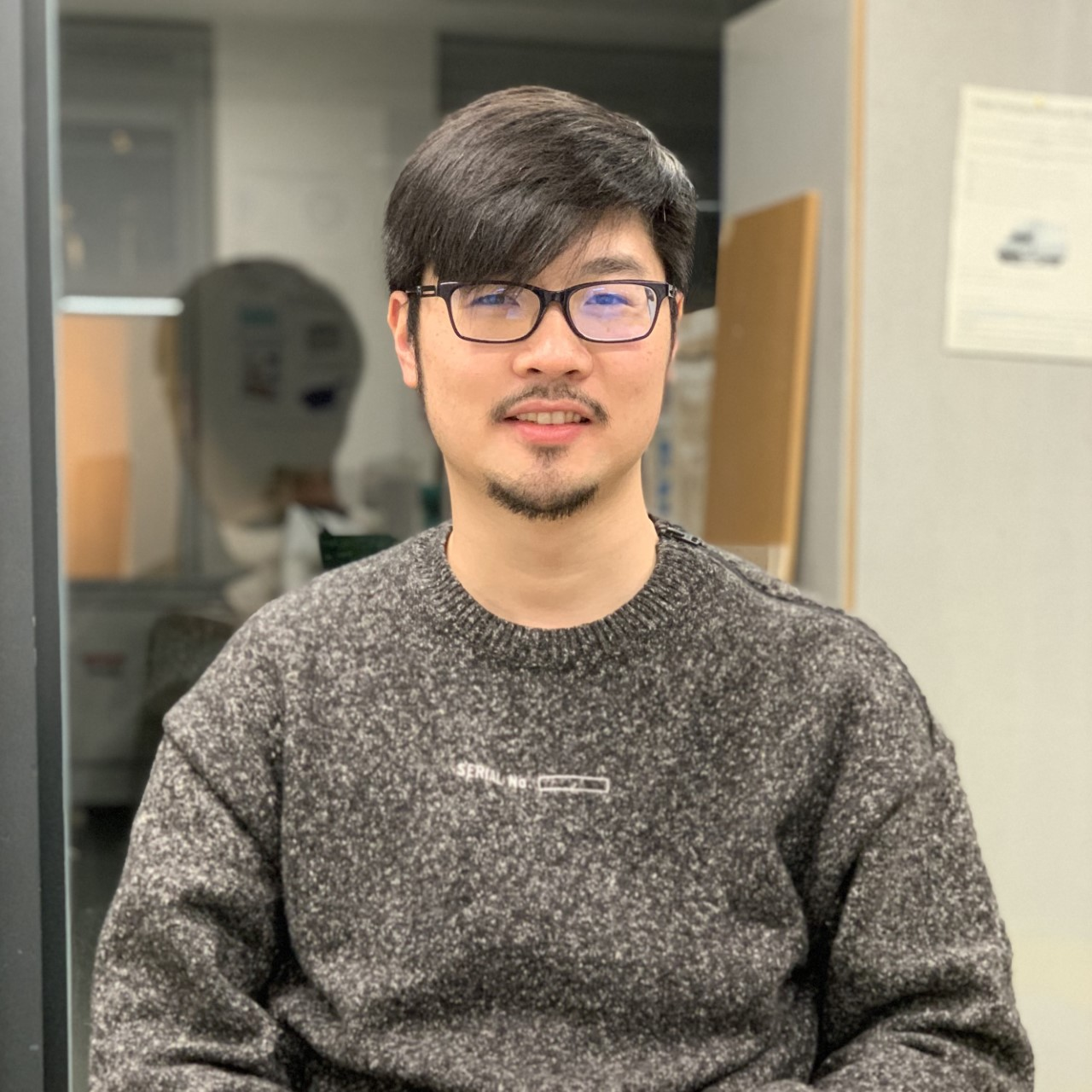}}]{Zhihua Wang} is a DPhil student at the Department of Computer Science, University of Oxford. Before that, he obtained MPhil degree from University of Cambridge and BEng degree from University of Manchester. His research interests lie in Cyber-Physical systems, deep learning and intuitive physics.
\end{IEEEbiography}

\begin{IEEEbiography}[{\includegraphics[width=1in,height=1in,clip]{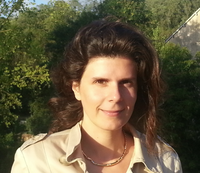}}]{Niki Trigoni} is a Professor at the Oxford University Department of Computer Science and a fellow of Kellogg College. She  obtained her DPhil at the University of Cambridge  (2001), became a postdoctoral researcher at Cornell University (2002-2004), and a Lecturer at Birkbeck College (2004-2007). At Oxford, she is currently  Director of the EPSRC Centre for Doctoral Training on  Autonomous Intelligent Machines and Systems, a program that combines machine learning, robotics, sensor systems and verification/control. She also leads the Cyber Physical Systems Group \url{http://www.cs.ox.ac.uk/activities/sensors/index.html},  which is focusing on intelligent and autonomous sensor  systems with applications in positioning, healthcare, environmental monitoring and smart cities. The groups  research ranges from novel sensor modalities and low  level signal processing to high level inference and learning.
\end{IEEEbiography}

\begin{IEEEbiography}[{\includegraphics[width=1in,height=1in,clip]{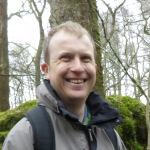}}]{Andrew Markham} is an Associate Professor at the Department of Computer Science, University of Oxford. He obtained his BSc  (2004) and PhD (2008) degrees from the University of Cape Town, South Africa. He is the Director of the MSc in Software Engineering. He works on resource-constrained systems, positioning systems, in particular magneto-inductive positioning and machine intelligence.
\end{IEEEbiography}
\clearpage
\begin{appendices}

\section{Details for the Evaluation of Sampling.}
We provide the implementation details of different sampling approaches evaluated in Section \ref{sec:eff_sampling}. To sample $M$ points (point features) from a large-scale point cloud $\boldsymbol{P}$ with $N$ points (point features):
\begin{enumerate}
\item \noindent\textit{Farthest Point Sampling (FPS):} We follow the implementation\footnote{\url{https://github.com/charlesq34/pointnet2}} provided by PointNet++ \cite{qi2017pointnet++}, which is also widely used in \cite{li2018pointcnn, wu2018pointconv, liu2019relation, chen2019lsanet, pointweb}. In particular, FPS is implemented as an operator running on GPU.
\item \noindent\textit{Inverse Density Importance Sampling (IDIS):} 
Given a point $p_i$, its density $\rho$ is approximated by calculating the summation of the distances between $p_i$ and its nearest $t$ points \cite{Groh2018flexconv}. Formally:
\begin{equation}
    \rho (p_{i})= \sum_{j=1}^{t} \left | \left | p_{i}-p_{i}^{j} \right | \right |, p_{i}^{j}\in \mathcal{N}(p_{i})
\end{equation}
\noindent where $p_{i}^{j}$ represents the coordinates (i.e. x-y-z) of the $j^{th}$ point of the neighbour points set $\mathcal{N}(p_{i})$, $t$ is set to 16. All the points are ranked according to the inverse density $\frac{1}{\rho}$ of points. Finally, the top $M$ points are selected.

\item\noindent\textit{Poisson Disk Sampling (PDS)}: We implement Fast Poisson Disk Sampling \cite{bridson2007fast} using the PDAL library\footnote{\url{https://pdal.io/}} for sampling $M$ points with Poisson disk property from an input cloud with $N$ points, given a Poisson radius $r$. We implement simple batch parallelisation. 

\item \noindent\textit{Random Sampling (RS):} We implement random sampling with the python numpy package. Specifically, we first use the numpy function \textit{numpy.random.choice()} to generate $M$ indices. We then gather the corresponding spatial coordinates and per-point features from point clouds by using these indices.

\item \noindent\textit{Generator-based Sampling (GS):} The implementation follows the code\footnote{\url{https://github.com/orendv/learning_to_sample}} provided by \cite{learning2sample}. We first train a ProgressiveNet \cite{learning2sample} to transform the raw point clouds into ordered point sets according to their relevance to the task. After that, the first $M$ points are kept, while the rest is discarded.

\item \noindent\textit{Continuous Relaxation based Sampling (CRS):}  \textit{CRS} is implemented with the self-attended gumbel-softmax sampling \cite{concrete}\cite{Yang2019ModelingPC}. Given a point feature set $\boldsymbol{P} \in \mathbb{R}^{N\times (d+3)}$ with 3D coordinates and per point features, we firstly estimate a probability score vector $\mathbf{s} \in \mathbb{R}^{N}$ through a score function parameterized by a MLP layer, i.e., $\mathbf{s}=softmax(MLP(\boldsymbol{P}))$, which learns a categorical distribution. Then, with the Gumbel noise $\mathbf{g} \in \mathbb{R}^{N}$ drawn from the distribution $Gumbel(0, 1)$. Each sampled point feature vector $\mathbf{y} \in \mathbb{R}^{d+3}$ is calculated as follows:
\begin{equation}\label{eq: concrete sampling}
    \mathbf{y} = \sum_{i=1}^N 
    \dfrac{\exp{((log (\mathbf{s}^{(i)})+\mathbf{g}^{(i)})/\tau)} \boldsymbol{P}^{(i)}}
    {\sum_{j=1}^N \exp{((log (\mathbf{s}^{(j)}) + \mathbf{g}^{(j)})/\tau)}},
\end{equation}
where $\mathbf{s}^{(i)}$ and $\mathbf{g}^{(i)}$ indicate the $i^{th}$ element in the vector $\mathbf{s}$ and $\mathbf{g}$ respectively, $\boldsymbol{P}^{(i)}$ represents the $i^{th}$ row vector in the input matrix $\boldsymbol{P}$. $\tau > 0$ is the annealing temperature. When $\tau \rightarrow 0$, Equation~\ref{eq: concrete sampling} approaches the discrete distribution and samples each row vector in $\boldsymbol{P}$ with the probability $p(\mathbf{y}=\boldsymbol{P}^{(i)})=\mathbf{s}^{(i)}$.

\item \noindent\textit{Policy Gradients based Sampling (PGS):} Given a point feature set $\boldsymbol{P} \in \mathbb{R}^{N\times (d+3)}$ with 3D coordinates and per point features, we first predict a score $\mathbf{s}$ for each point, which is learnt by an MLP function, i.e., $\mathbf{s}=softmax(MLP(\boldsymbol{P}))+\mathbf{\epsilon}$, where $\mathbf{\epsilon} \in \mathbb{R}^N$ is a zero-mean Gaussian noise with the variance $\mathbf{\Sigma}$ for random exploration. After that, we sample $M$ vectors in $\boldsymbol{P}$ with the top $M$ scores. Sampling each point/vector can be regarded as an independent action and a sequence of them form a sequential Markov Decision Process (MDP) with the following policy function $\pi$:
\begin{equation}
    a_i\sim \pi(a|\boldsymbol{P}^{(i)}; \theta, \mathbf{s})
\end{equation}
where $a_i$ is the binary decision of whether to sample the $i^{th}$ vector in $\boldsymbol{P}$ and $\theta$ is the network parameter of the MLP.
Hence to properly improve the sampling policy with an underivable sampling process, we apply REINFORCE algorithm \cite{sutton2000policy} as the gradient estimator. The segmentation accuracy $R$ is applied as the reward value for the entire sampling process as $\mathcal{J}=R$. It is optimized with the following estimated gradients:
\begin{equation}\label{eq: reinforce}
  \begin{aligned}
    \dfrac{\partial \mathcal{J} }{\partial \theta} \approx  \dfrac{1}{M}\sum_{m=1}^M\sum_{i=1}^{N} \dfrac{\partial}{\partial \theta}\log \pi(a_i|\boldsymbol{P}^{(i)};\theta,\mathbf{\Sigma}) \times \\
    (R-b^c-b(\boldsymbol{P}^{(i)})),
    \end{aligned}
\end{equation}
where $M$ is the batch size, $b^c$ and $b(\boldsymbol{P}^{(i)})$ are two control variates \cite{mnih2014neural} for alleviating the high variance problem of policy gradients.
\end{enumerate}

\section{Additional experimental results.}
\begin{table*}[thb]
\centering
\caption{Quantitative results of different approaches on ScanNet (online test set). }
\label{tab:scannet}
\resizebox{\textwidth}{!}{%
\begin{tabular}{rccccccccccccccccccccc}
\Xhline{2\arrayrulewidth}
Method & mIoU(\%) & floor & wall & chair & sofa & table & door & cab & bed & desk & toil & sink & wind & pic & bkshf & curt & show & cntr & fridg & bath & other \\
\Xhline{2\arrayrulewidth}
ScanNet \cite{scannet} & 30.6 & 78.6 & 43.7 & 52.4 & 34.8 & 30.0 & 18.9 & 31.1 & 36.6 & 34.2 & 46.0 & 31.8 & 18.2 & 10.2 & 50.1 & 0.2 & 15.2 & 21.1 & 24.5 & 20.3 & 14.5 \\
PointNet++ \cite{qi2017pointnet++} & 33.9 & 67.7 & 52.3 & 36.0 & 34.6 & 23.2 & 26.1 & 25.6 & 47.8 & 27.8 & 54.8 & 36.4 & 25.2 & 11.7 & 45.8 & 24.7 & 14.5 & 25.0 & 21.2 & 58.4 & 18.3 \\
SPLATNET3D \cite{su2018splatnet} & 39.3 & 92.7 & 69.9 & 65.6 & 51.0 & 38.3 & 19.7 & 31.1 & 51.1 & 32.8 & 59.3 & 27.1 & 26.7 & 0.0 & 60.6 & 40.5 & 24.9 & 24.5 & 0.1 & 47.2 & 22.7 \\
Tangent-Conv \cite{tangentconv} & 43.8 & 91.8 & 63.3 & 64.5 & 56.2 & 42.7 & 27.9 & 36.9 & 64.6 & 28.2 & 61.9 & 48.7 & 35.2 & 14.7 & 47.4 & 25.8 & 29.4 & 35.3 & 28.3 & 43.7 & 29.8 \\
PointCNN \cite{li2018pointcnn} & 45.8 & 94.4 & 70.9 & 71.5 & 54.5 & 45.6 & 31.9 & 32.1 & 61.1 & 32.8 & 75.5 & 48.4 & 47.5 & 16.4 & 35.6 & 37.6 & 22.9 & 29.9 & 21.6 & 57.7 & 28.5 \\
PointConv \cite{wu2018pointconv} & 55.6 & 94.4 & 76.2 & 73.9 & 63.9 & 50.5 & 44.5 & 47.2 & 64.0 & 41.8 & 82.7 & 54.0 & 51.5 & 18.5 & 57.4 & 43.3 & 57.5 & 43.0 & 46.4 & 63.6 & 37.2 \\
SPH3D-GCN \cite{SPH3D-GCN} & 61.0 & 93.5 & 77.3 & 79.2 & 70.5 & 54.9 & 50.7 & 53.2 & 77.2 & 57.0 & 85.9 & 60.2 & 53.4 & 4.6 & 48.9 & 64.3 & 70.2 & 40.4 & 51.0 & 85.8 & 41.4 \\
KPConv \cite{thomas2019kpconv} & 68.4 & 93.5 & 81.9 & 81.4 & 78.5 & 61.4 & 59.4 & 64.7 & 75.8 & 60.5 & 88.2 & 69.0 & 63.2 & 18.1 & 78.4 & 77.2 & 80.5 & 47.3 & 58.7 & 84.7 & 45.0 \\
SparseConvNet \cite{sparse} & 72.5 & 95.5 & 86.5 & 86.9 & 82.3 & 62.8 & 61.4 & 72.1 & 82.1 & 60.3 & 93.4 & 72.4 & 68.3 & 32.5 & 84.6 & 75.4 & 87.0 & 53.3 & 71.0 & 64.7 & 57.2 \\
SegGCN \cite{lei2020seggcn} & 58.9 & 93.6 & 77.1 & 78.9 & 70.0 & 56.3 & 48.4 & 51.4 & 73.1 & 57.3 & 87.4 & 59.4 & 49.3 & 6.1 & 53.9 & 46.7 & 50.7 & 44.8 & 50.1 & 83.3 & 39.6 \\
\textbf{RandLA-Net (Ours)} & 64.5     & 94.5  & 79.2 & 82.9  & 73.8 & 59.9  & 52.3 & 57.7 & 73.1 & 47.7 & 82.7 & 61.8 & 62.1 & 26.9 & 69.9  & 73.6 & 74.9 & 44.6 & 48.4  & 77.8 & 45.4 \\
\Xhline{2\arrayrulewidth}
\end{tabular}%
}
\end{table*}

\begin{table*}[thb]
\begin{center}
\caption{Quantitative results of different approaches on the DALES dataset. Overall Accuracy (OA, \%), mean class Accuracy (mAcc, \%), mean IoU (mIoU, \%), and per-class IoU (\%) are reported.}
\resizebox{0.95\textwidth}{!}{%
\begin{tabular}{rcccccccccc}
\Xhline{2\arrayrulewidth}
Method & OA(\%) & mIoU(\%) & \textit{ground}& \textit{buildings}&\textit{cars}&\textit{trucks}&\textit{poles}&\textit{power lines}&\textit{fences}&\textit{vegetation}      \\
\Xhline{2\arrayrulewidth}
ShellNet \cite{zhang2019shellnet} & 96.4&	57.4& 96.0 &	95.4&	32.2&	39.6&	20.0&	27.4&	60.0&	88.4\\
PointCNN \cite{li2018pointcnn} & 97.2	&58.4&	\textbf{97.5}&	95.7&	40.6&	4.80&	57.6&	26.7&	52.6&	91.7\\
SuperPoint \cite{landrieu2018large} &95.5&	60.6&	94.7&	93.4&	62.9&	18.7&	28.5&	65.2&33.6&	87.9\\
ConvPoint \cite{conv_pts} &97.2&	67.4&	96.9&	96.3&	75.5&	21.7&	40.3&	86.7&	29.6&	91.9\\
PointNet++ \cite{qi2017pointnet++} & 95.7	&68.3	&94.1&	89.1&	75.4&	30.3&	40.0& 79.9&	46.2&	91.2\\
KPConv \cite{thomas2019kpconv}& \textbf{97.8}&	\textbf{81.1}&	97.1&	\textbf{96.6}&	\textbf{85.3}&	41.9&	\textbf{75.0}&	\textbf{95.5}&	63.5&	94.1\\
\textbf{\nickname{} (Ours)} & 97.1	&80.0	&97.0&	93.2&	83.7&	\textbf{43.8}&	59.4& 94.8&	\textbf{71.5}&	\textbf{96.6}\\
\Xhline{2\arrayrulewidth}
\end{tabular}}
\label{tab:dales}
\end{center}
\end{table*}

\subsection{Evaluation on ScanNet}

We conduct an experiment on the ScanNet dataset to further evaluate the performance of our method. The detailed results on the online test set are shown in Table \ref{tab:scannet}.

It can be seen that our approach achieves satisfactory results on this dataset, but not the best. Our mean IoU score is lower than  KPConv \cite{thomas2019kpconv} and SparseConvNet \cite{sparse}, because our RandLA-Net performs much worse on several small categories such as \textit{refrigerator} and \textit{cabinet}. One possible reason is that the major information of small objects may be lost due to the aggressive random sampling used in our framework. Note that, the ScanNet dataset is mainly composed of relatively small indoor scenes (average spatial size:               5m$\times$5m$\times$2m\footnote{This is computed by averaging the spatial sizes of all scenes in the dataset.}), hence it is not in favor of our method which is primarily designed for large-scale point clouds (e.g., point clouds spanning up to 250m$\times$260m$\times$80m in Semantic3D dataset). \revminor{We observe that the performance gap between our approach and KPConv on the ScanNet is greater than that on the S3DIS dataset. This may be caused by the KNN used in our framework, where the receptive field of each point is not stable as that of the spherical neighborhood used in KPConv, especially for the RGBD-reconstructed dataset such as ScanNet (extremely non-uniform, incomplete, and noisy.)}

\subsection{Evaluation on DALES}
The DALES dataset \cite{varney2020dales} consists of large-scale point clouds with 505 million labeled points acquired by the aerial laser systems (ALS), which covers urban areas spanning 10 $km^2$. Each point is labeled with one of 8 semantic categories, and there is no color information available in this dataset.

Following the evaluation protocol in \cite{varney2020dales}, we further compare the segmentation performance of our RandLA-Net with other approaches on this dataset. The detailed results are shown in Table \ref{tab:dales}. It can be seen that our RandLA-Net also achieves satisfactory overall performance on this dataset, with the mean IoU score of 80.0\%. It also achieves the best performance on categories such as \textit{trucks} and \textit{vegetation}. We also notice that the performance is slightly lower than KPConv \cite{thomas2019kpconv}, primarily because of the poor performance on the \textit{poles}. This is likely caused by the aggressive downsampling used in our framework.

\subsection{Ablation of Voting Scheme}

As presented in Section \ref{subsec:Implementation}, the voting scheme is applied to obtain the final predictions during inference. To further investigate the impact of the voting scheme on the overall segmentation performance, we conduct the following experiments by removing the voting scheme from our RandLA-Net. We obtained new mIoU scores on Toronto-3D \cite{Toronto3D}, NPM3D \cite{NPM3D}, and SemanticKITTI \cite{behley2019semantickitti} datasets. In particular, instead of voting from multiple runs, we only keep the per-point result in the last inference. The detailed results are shown in Table \ref{tab:voting}. 

\begin{table}[thb]
\centering
\caption{The mIoU scores (\%) of our RandLA-Net with and without the voting scheme.}
\label{tab:voting}
\resizebox{0.45\textwidth}{!}{%
\begin{tabular}{rccc}
\Xhline{2\arrayrulewidth}
& \begin{tabular}[c]{@{}c@{}}Toronto-3D \cite{Toronto3D}\\ (Test set)\end{tabular} & \begin{tabular}[c]{@{}c@{}}NPM3D \cite{NPM3D}\\ (Test set)\end{tabular} & \begin{tabular}[c]{@{}c@{}}SemanticKITTI \cite{behley2019semantickitti}\\ (Validation set)\end{tabular} \\ 
\Xhline{2\arrayrulewidth}
W/O voting           &  79.4           & 76.8      & 53.0       \\
W/ voting            &  81.8          & 78.8      & 57.1      \\        \Xhline{2\arrayrulewidth}
\end{tabular}%
}
\end{table}

\begin{table}[thb]
\centering
\caption{Quantitative results of the proposed RandLA-Net on the SemanticKITTI dataset (validation set) with different upsampling strategies.}
\label{tab:upsampling}
\resizebox{0.35\textwidth}{!}{%
\begin{tabular}{rc}
\Xhline{2\arrayrulewidth}
Methods                   & mIoU(\%) \\
\Xhline{2\arrayrulewidth}
Random-upsampling         & 29.2     \\
Trilinear-upsampling     & 53.0     \\
Attentive upsampling (K=3) & 56.6     \\
Attentive upsampling (K=5) & 57.0     \\
Attentive upsampling (K=8) & 56.1     \\
Nearest-upsampling        & 57.1    \\
\Xhline{2\arrayrulewidth}
\end{tabular}
}
\end{table}

It can be seen that the overall mIoU scores of RandLA-Net has decreased by 2\%-4\% on Toronto-3D \cite{Toronto3D}, NPM3D \cite{NPM3D}, and SemanticKITTI \cite{behley2019semantickitti} datasets, showing that the commonly used voting scheme is helpful to boost the performance, yet it may incur extra computation burdens.

\begin{table*}[thb]
\centering
\caption{Quantitative results of the proposed RandLA-Net on the S3DIS dataset (Area 5) with different preprocessing steps.}
\label{tab:preprocessing}
\resizebox{\textwidth}{!}{%
\begin{tabular}{rccccccccccccccc}
\Xhline{2\arrayrulewidth}
Method & Size & mIoU(\%) & ceil. & floor & wall & beam & col. & wind. & door & table & chair & sofa & book. & board & clut. \\
\Xhline{2\arrayrulewidth}
Grid (grid size=0.04m) & 40960 pts & 62.67 & 92.41 & 97.09 & 81.03 & 0.00 & 28.74 & 62.62 & 37.87 & 78.70 & 85.45 & 66.83 & 69.59 & 63.72 & 50.64 \\
Block random (1m$\times$1m) & 4096 pts & 59.10 & 90.27 & 96.53 & 77.82 & 0.00 & 30.32 & 65.28 & 36.10 & 72.02 & 79.82 & 55.00 & 64.48 & 56.30 & 44.33 \\
Block random (6m$\times$6m) & 40960 pts & 61.63 & 88.60 & 97.30 & 80.88 & 0.00 & 35.16 & 62.04 & 45.92 & 76.30 & 81.97 & 65.10 & 67.92 & 53.05 & 46.91 \\
\Xhline{2\arrayrulewidth}
\end{tabular}%
}
\end{table*}

\begin{table*}[thb]
\centering
\caption{Quantitative results of the proposed RandLA-Net on the SemanticKITTI dataset (validation set) with deeper layers and wider channels.}
\label{tab:deeper_wider}
\resizebox{\textwidth}{!}{%
\begin{tabular}{rlllllllllllllllllllll}
\Xhline{2\arrayrulewidth}
Methods & \rotatebox{90}{Params(M)} & \rotatebox{90}{\textbf{mIoU(\%)}} & \rotatebox{90}{road} & \rotatebox{90}{sidewalk} & \rotatebox{90}{parking} & \rotatebox{90}{other-ground} & \rotatebox{90}{building} & \rotatebox{90}{car} & \rotatebox{90}{truck} & \rotatebox{90}{bicycle} & \rotatebox{90}{motorcycle} & \rotatebox{90}{other-vehicle} & \rotatebox{90}{vegetation} & \rotatebox{90}{trunk} & \rotatebox{90}{terrain} & \rotatebox{90}{person} & \rotatebox{90}{bicyclist} & \rotatebox{90}{motorcyclist} & \rotatebox{90}{fence} & \rotatebox{90}{pole} & \rotatebox{90}{traffic-sign} \\
\Xhline{2\arrayrulewidth}
\nickname{} (Deeper) & 20.0  & 56.3 & 93.4 & 20.0 & 34.2 & 84.0 & 41.1 & 50.9 & 64.9 & 0.00 & 92.7 & 42.3 & 79.7 & 0.09 & 89.3 & 55.4 & 86.8 & 63.5 & 76.6 & 55.5 & 38.9 \\
\nickname{} (Wider) & 4.9 & 57.2 & 94.8 & 19.9 & 35.2 & 68.6 & 48.0 & 56.5 & 71.2 & 0.00 & 93.5 & 45.3 & 80.8 & 1.80 & 89.7 & 55.9 & 86.7 & 66.0 & 74.3 & 57.8 & 41.1 \\
\textbf{\nickname{} (Standard)} & 1.2 & 57.1 & 94.2 & 16.7 & 42.9 & 71.9 & 40.3 & 57.3 & 71.5 & 0.00 & 92.7 & 42.4 & 80.0 & 3.25 & 89.3 & 56.0 & 86.6 & 65.5 & 74.6 & 57.5 & 40.7 \\

\Xhline{2\arrayrulewidth}
\end{tabular}%
}
\end{table*}

\subsection{Ablation of Nearest-Neighbor Interpolation}
As mentioned in Sec. \ref{network_structure}, we choose the nearest-neighbor interpolation for simplicity and efficiency. However, our framework is flexible and allows the integration of different weighted interpolation strategies or other learnable layers.  Therefore, we further compare the impact of different upsampling strategies on the overall segmentation performance, including random upsampling and weighted interpolation (e.g., trilinear interpolation). In particular, we implement random upsampling by randomly copy the point features from the current feature map to create the upsampled feature map. For trilinear interpolation, we follow \cite{qi2017pointnet++} to weight the point features of the nearest three points according to the Euclidean distance. The detailed results are shown in Table \ref{tab:upsampling}.

It can be seen that, (1) random upsampling is not applicable in the decoding stage, primarily because the skip connection used in our framework requires that the features at the same spatial location should be related, instead of being randomly selected. (2) the mIoU scores with trilinear interpolation and attentive upsampling are inferior to the nearest upsampling (57.1\%). One possible reason is that the input point clouds are significantly down-sampled after all encoding layers, which means that the final subset of points preserved are very sparse and have different semantic meanings. Therefore, at the beginning stage of up-sampling, if multiple points with different semantic meanings are selected (e.g., by KNN) to trilinearly interpolate the features of a new point, it might be detrimental. Instead, if only the nearest point is selected, it tends to be more robust.

\subsection{Ablation of Preprocessing}
Different preprocessing steps are adopted for different segmentation approaches. For example, PointNet-based methods \cite{qi2017pointnet, qi2017pointnet++, wu2018pointconv} adopt block random sampling (e.g., sampling 4k points from 1mx1m blocks) to preprocess the point clouds at the beginning, while KPConv \cite{thomas2019kpconv} and SCN \cite{sparse} first use grid sampling to process the point clouds. Here,  we conduct additional experiments to further compare the segmentation performance of our method under different preprocessing steps. In particular, we evaluate the performance of our framework with grid sampling, block random sampling (1m$\times$1m with 4096 points and 6m$\times$6m with 40960 points). The detailed results are shown in Table \ref{tab:preprocessing}.

It can be seen that our framework achieves comparable performance under different preprocessing steps when being fed with large point clouds (i.e., grid subsampling with 40960 points vs. block random sampling (6m$\times$6m with 40960 points)). However, the segmentation performance decreases from 61.63\% to 59.10\% when the block size is reduced to 1m$\times$1m, because the excessive block partitions severely break the geometrical structures and the network is hard to learn meaningful and robust representations. For the detailed analysis of the data preparation, please refer to \cite{hu2020towards}.

\begin{table*}[thb]
\centering
\caption{Quantitative results of the proposed RandLA-Net on the S3DIS dataset (Area 5) with 5 runs. Overall Accuracy (OA, \%), mean class Accuracy (mAcc, \%), mean IoU (mIoU, \%), and per-class IoU (\%) are reported. }
\label{tab:s3dis-5-runs}
\resizebox{\textwidth}{!}{%
\begin{tabular}{rcccccccccccccc}
\Xhline{2\arrayrulewidth}
& mIoU(\%) & ceil. & floor & wall & beam & col. & wind. & door & table & chair & sofa & book. & board & clut. \\
\Xhline{2\arrayrulewidth}
Iter1 & 63.75 & 92.19 & 97.67 & 81.12 & 0.00 & 20.22 & 61.02 & 41.49 & 78.53 & 88.04 & 74.21 & 70.65 & 70.65 & 53.01 \\
Iter2 & 61.56 & 92.51 & 96.18 & 80.92 & 0.00 & 20.00 & 60.63 & 37.93 & 76.37 & 87.68 & 58.63 & 70.33 & 67.13 & 51.94 \\
Iter3 & 61.94 & 92.05 & 97.24 & 80.95 & 0.00 & 13.81 & 61.62 & 36.38 & 77.79 & 87.08 & 65.81 & 71.02 & 68.41 & 53.06 \\
Iter4 & 62.30 & 92.28 & 97.59 & 80.37 & 0.00 & 23.11 & 60.99 & 33.32 & 80.01 & 87.42 & 63.46 & 70.42 & 68.45 & 52.45 \\
Iter5 & 62.67 & 92.41 & 97.09 & 81.03 & 0.00 & 28.74 & 62.62 & 37.87 & 78.70 & 85.45 & 66.83 & 69.59 & 63.72 & 50.64 \\
Average & 62.44 & 92.29 & 97.15 & 80.88 & 0.00 & 21.18 & 61.38 & 37.40 & 78.28 & 87.13 & 65.79 & 70.40 & 67.67 & 52.22 \\
STD & 0.75 & 0.16 & 0.53 & 0.26 & 0.00 & 4.85 & 0.70 & 2.64 & 1.19 & 0.90 & 5.07 & 0.47 & 2.28 & 0.89 \\
\Xhline{2\arrayrulewidth}
\end{tabular}%
}
\end{table*}

\subsection{Sensitivity of \nickname{}}
Due to the usage of random sampling, the segmentation results of our RandLA-Net may be different given different runs. To further evaluate the sensitivity (e.g., variance) of our method, we run the well trained model on the S3DIS dataset (Area 5) 5 times, with exactly the same experimental settings. The detailed results are shown in Table \ref{tab:s3dis-5-runs}.

The average mIoU scores of our method on the Area5 of the S3DIS dataset is 62.44\%, with a standard deviation of 0.75. We observe that the main difference of results lies in the minor categories such as \textit{sofa}, \textit{column}, \textit{board} and \textit{door}. Since these categories only have a small number of points, their segmentation results have noticeable fluctuations within multiple runs.

\subsection{Variants of RandLA-Net}
RandLA-Net can be deeper with more layers and wider with more channels. However, deeper or wider networks inevitably introduce larger models and massive training parameters, making the optimization more difficult in practice. Further, the overall computation efficiency would also be sacrificed.

In this section, we compare the segmentation performance of RandLA-Net with deeper layers and wider channels. In particular, for a deeper model, we increase the number of encoding layers from 4 to 8, and the feature dimensions gradually increase from $(8\rightarrow16\rightarrow32\rightarrow64\rightarrow 128\rightarrow 256\rightarrow 512\rightarrow1024)$. Accordingly, the downsampling rate is decreased from 4 to 2 to preserve sufficient information for the decoder. For a wider model, we double the predefined per-point feature dimensions in our framework, that is: $(16\rightarrow64\rightarrow 256\rightarrow 512\rightarrow1024)$. All other hyperparameters such as the number of nearest neighbors, learning rate, batch size, and the number of points per batch remain unchanged for a fair comparison. The detailed results achieved on the SemanticKITTI dataset are shown in Table \ref{tab:deeper_wider}.

It can be seen that the segmentation performance is slightly improved when more channels are added (i.e., the wider model), but slightly decreased when more layers are added (i.e, the deeper model). One possible reason is that the deeper model has nearly 20 times more parameters, making it prone to overfitting and difficult to optimize. Additionally, the same hyperparameters may not always be optimal for each model.

\subsection{Overfitting issue of RandLA-Net}
Several point-based methods \cite{thomas2019kpconv, li2018pointcnn, wu2018pointconv} tend to be overfitting training datasets. We present both the training and validation accuracy of our RandLA-Net in three datasets in Table \ref{tab:overfitting}.

\begin{table}[h]
\centering
\caption{Training and validation accuracy of our RandLA-Net on three datasets.}
\label{tab:overfitting}
\resizebox{0.4\textwidth}{!}{%
\begin{tabular}{rcc}
\Xhline{2\arrayrulewidth}
\multirow{2}{*}{Datasets} & \multicolumn{2}{c}{Overall Accuracy (\%)} \\
                          & Training Set       & Validation set       \\
\Xhline{2\arrayrulewidth}
Toronto3D \cite{Toronto3D}                & 93.6                  & 93.0                \\
S3DIS (Area5) \cite{2D-3D-S}                    & 95.7                  & 87.1                 \\
SemanticKITTI \cite{behley2019semantickitti}            & 94.3               & 90.9                \\
\Xhline{2\arrayrulewidth}
\end{tabular}}
\end{table}

Overall, our RandLA-Net has good generalization ability, without suffering from overfitting. The reasons are two-fold. First, RandLA-Net is lightweight with far fewer network parameters (1.24M) than PointConv \cite{wu2018pointconv} (21.7M), KPConv \cite{thomas2019kpconv} (14.3M), and PointCNN \cite{li2018pointcnn} (11.5M). These parameters are shared by a large amount of 3D points and their neighbours, thus capturing more general patterns. Second, the random sampling in our framework can be regarded as an implicit way of data augmentation and regularization. With the same input point cloud, the preserved 3D points can be different in multiple training iterations, thereby preventing the network from overfitting the training set.

\end{appendices}

\end{document}